\documentclass{article}

\usepackage[preprint,nonatbib]{neurips_2020}

\usepackage{color}
\definecolor{bleudefrance}{rgb}{0.19, 0.55, 0.91}
	
\usepackage[utf8]{inputenc} % allow utf-8 input
\usepackage[T1]{fontenc}    % use 8-bit T1 fonts
\usepackage[colorlinks=True, citecolor=bleudefrance, linkcolor=bleudefrance]{hyperref}       % hyperlinks
\usepackage{url}            % simple URL typesetting
\usepackage{booktabs}       % professional-quality tables
\usepackage{amsfonts}       % blackboard math symbols
\usepackage{nicefrac}       % compact symbols for 1/2, etc.
\usepackage{microtype}      % microtypography
\usepackage{graphicx}
\usepackage{amsmath,amssymb}
\usepackage{multirow}
\usepackage{enumitem}

\newcounter{descriptcount}
\renewcommand*\thedescriptcount{\arabic{descriptcount}}

\usepackage{array}

\usepackage{wrapfig}
\usepackage{svg}

\DeclareMathOperator{\Ex}{\mathbb{E}}
\DeclareMathOperator{\loss}{\mathcal{L}}
\DeclareMathOperator{\dist}{\mathcal{D}}
\newcommand{\standardloss}[2]{\min\limits_\theta \Ex_{(#1, #2) \sim \dist} [\loss_\theta (#1, #2)]}
\newcommand{\adversarialloss}[2]{\min\limits_\theta \Ex_{(#1, #2) \sim \dist} \left[ \max\limits_{\delta \in \Delta(#1)} \loss_\theta (#1 + \delta, #2)\right]}

\newcolumntype{H}{>{\setbox0=\hbox\bgroup}c<{\egroup}@{}}

\title{Representation Quality Of Neural Networks\\ Links To Adversarial Attacks and Defences}

\author{%
  Shashank Kotyan\\
  Kyushu University \\
  \texttt{shashankkotyan@gmail.com}\\
  \And
  Danilo Vasconcellos Vargas\\
  Kyushu University \\
  \texttt{vargas@inf.kyushu-u.ac.jp}\\
  \And
  Moe Matsuki \\
  Kyushu Institute of Technology \\
  \texttt{matsuki.sousisu@gmail.com}\\
}

\begin{document}

\maketitle

% \newlength{\oldintextsep}
% \setlength{\oldintextsep}{\intextsep}
% \setlength\intextsep{0pt}

% \newlength{\oldtextfloatsep}
% \setlength{\oldtextfloatsep}{\textfloatsep}
% \setlength\textfloatsep{8pt}

\begin{abstract}

    Neural networks have been shown vulnerable to a variety of adversarial algorithms. 
    A crucial step to understanding the rationale for this lack of robustness is to assess the potential of the neural networks' representation to encode the existing features. 
    Here, we propose a method to understand the representation quality of the neural networks using a novel test based on Zero-Shot Learning, entitled Raw Zero-Shot. 
    The principal idea is that, if an algorithm learns rich features, such features should be able to interpret "\textit{unknown}" classes as an aggregate of previously learned features.
    This is because unknown classes usually share several regular features with recognised classes, given the features learned are general enough. 
    We further introduce two metrics to assess these learned features to interpret unknown classes.  
    One is based on inter-cluster validation technique (Davies-Bouldin Index), and the other is based on the distance to an approximated ground-truth. 
    Experiments suggest that adversarial defences improve the representation of the classifiers, further suggesting that to improve the robustness of the classifiers, one has to improve the representation quality also.
    Experiments also reveal a strong association (a high Pearson Correlation and low p-value) between the metrics and adversarial attacks. 
    Interestingly, the results indicate that dynamic routing networks such as CapsNet have better representation while current deeper neural networks are trading off representation quality for accuracy. 
    % This might serve as a guide to develop better models as well as be used in loss functions to formulate new types of neural networks.

\end{abstract}

\section{Introduction}

    Adversarial samples are noise-perturbed samples that can fail neural networks for tasks like image classification.
    Since they were discovered some years ago in \cite{szegedy2014intriguing}, the quality and the variety of adversarial samples have grown.
    These adversarial samples can be generated by a specific class of algorithms known as adversarial attacks \cite{nguyen2015deep,brown2017adversarial,moosavi2017universal,su2019one}.
    Most of these adversarial attacks can be transformed into real-world attacks \cite{sharif2016accessorize,kurakin2016adversarial,athalye2017synthesizing}, which confers a big issue as well as a security risk for current neural networks' applications.
    Albeit the existence of many defences \cite{goodfellow2014explaining,huang2015learning,papernot2016distillation,dziugaite2016study,hazan2016perturbations,das2017keeping,guo2017countering,song2017pixeldefend,xu2017feature,madry2017towards,ma2018characterizing,buckman2018thermometer} to these adversarial attacks, no known learning algorithm or procedure can defend consistently \cite{carlini2017towards,tramer2017ensemble,athalye2018obfuscated,uesato2018adversarial,vargas2019robustness}.
    This shows that a more profound understanding of the adversarial algorithms is needed to enable the formulation of consistent defences. 

    Few works have focused on understanding the reasoning behind such a lack of robustness.
    It is discussed in \cite{goodfellow2014explaining}, that neural networks' linearity is one of the main reasons for failure.
    Other investigation by \cite{thesing2019ai} shows that with deep learning, neural networks learn false structures that are simpler to learn rather than the ones expected. 
    Moreover, recent research by \cite{vargas2019understanding} unveil that adversarial attacks are altering where the algorithm is paying attention.
    We try to open up a new perspective on understanding adversarial algorithms based on representation quality. 
    We do this, by verifying that the representation quality of neural networks is indeed linked with the adversarial algorithms.

    \begin{wrapfigure}[16]{R}{0.5\linewidth}
        \centering
        \includegraphics[width=\linewidth]{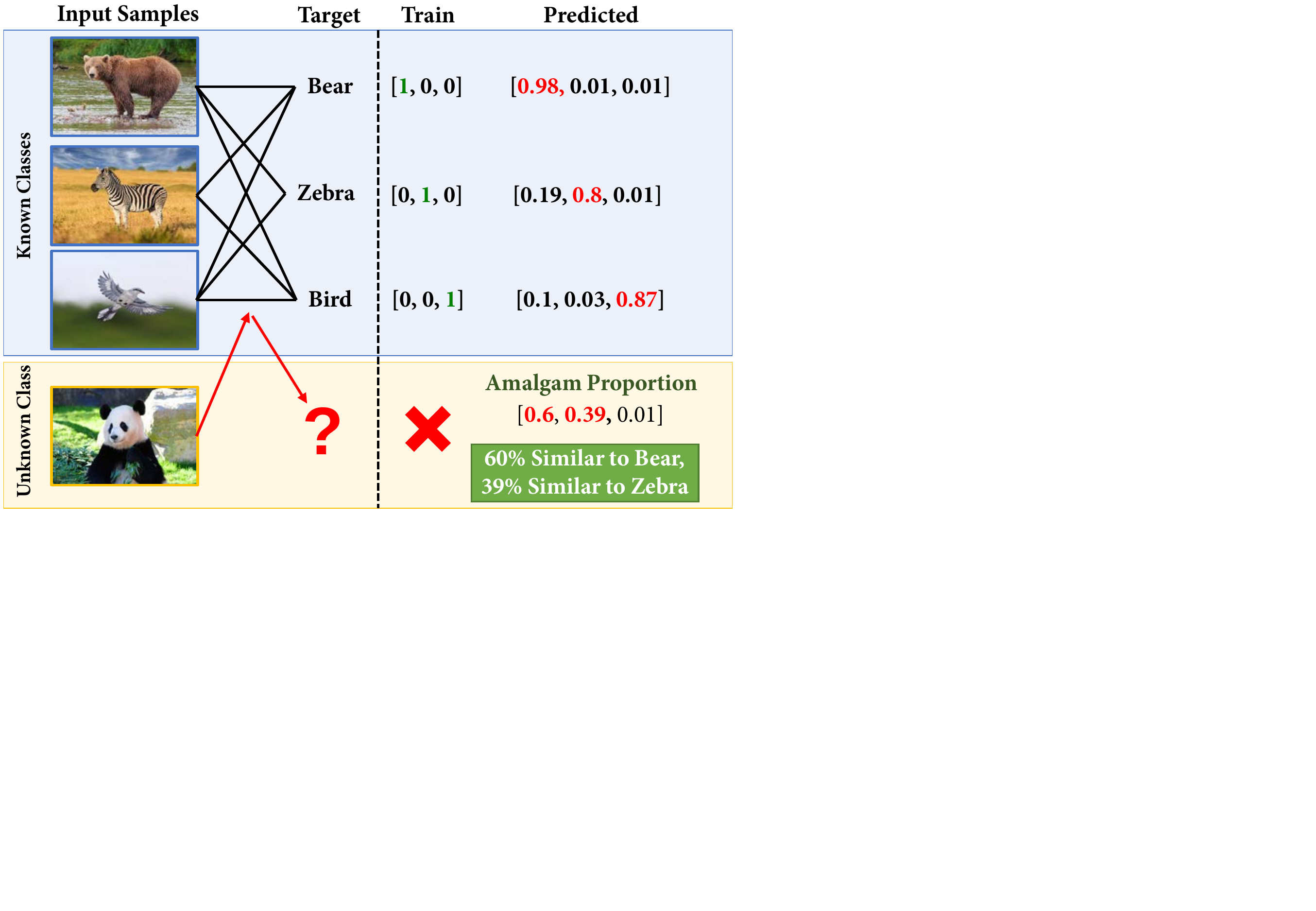} 
        \caption{Raw Zero-Shot Illustration}
        \label{figure_overview} 
    \end{wrapfigure}

    \paragraph*{Contributions:} In this article, we propose a methodology based on Zero-Shot Learning for evaluating the representation quality of the neural networks.
    We conduct the experiments over the soft-labels of an unfamiliar class to assess the representation quality of the classifiers.
    This is based on the hypothesis that if a classifier is capable of learning useful features, an unfamiliar class would also be associated with some of these learned features (Figure \ref{figure_overview}).
    We call this type of inspection over unfamiliar class, Raw Zero-Shot (Section \ref{section_raw_zero}).
    Furthermore, we also introduce two associated metrics to evaluate the representation quality of neural networks.
    One is based on Clustering Hypothesis (Section \ref{section_cluster_hypothesis}), while the other is based on Amalgam Hypothesis (Section \ref{section_amalgam_hypothesis}).

    We evaluate our Raw Zero-Shot test and our two metrics over a wide assortment of datasets (and classifiers) such as Fashion MNIST, CIFAR-10, and a customised Imagenet to assess the representation quality of the vanilla classifiers (Section \ref{section_results}).
    We also evaluate different adversarial defences to prove that these adversarial defences when applied to a classifier give better representation quality than the vanilla classifier (Section \ref{section_results}).
    Based on our Raw Zero-Shot test and our two metrics, we then reveal a link between the representation quality and attack susceptibility by verifying that the proposed metrics have a high Pearson Correlation with the adversarial attacks (Section \ref{section_link_attack}). 

\section{Related Works}

    \begin{description}[leftmargin=*]
    \item[Understanding Adversarial Attacks:] 
    Since the discovery of adversarial samples \cite{szegedy2014intriguing}, many researchers have tried to understand the adversarial attacks. 
    It is hypothesised that neural networks' linearity is one of the principal reasons for failure against an adversary \cite{goodfellow2014explaining} and 
    non-linear neural networks are thus, more robust compared to linear networks \cite{guo2018sparse}. 
    Based on this understanding, the authors of \cite{buckman2018thermometer} proposed to discretise the input feature space, which may lead to breaking this linearity.
    However, recent research proves that the input feature space itself is vast, which provide opportunities to the adversaries \cite{xu2017feature}.
    It was also observed that the classifiers are not familiarised with the "adversarial" input feature space as adversarial samples have much lower probability densities under the image distribution \cite{song2017pixeldefend}. 
    However, researchers also argue that adversarial attacks are entangled with interpretability of neural networks as results on adversarial samples can hardly be explained \cite{tao2018attacks}.
    Intuitively thus, in \cite{das2017keeping} authors recommended discarding some of the information unnoticeable to humans in input feature space by compressing as adversarial noises are often indiscernible by the human eye.
    % The bounds for the robustness using this input feature space is also studied in \cite{fawzi2018adversarial}.
    Another aspect of robustness is discussed in \cite{madry2017towards}, where authors suggest that the capacity of the neural networks' architecture is relevant to the robustness.
    In this article, we explore a new perspective to understand adversarial attacks and defences based on the representation quality of the neural networks.
    % \cite{huang2015learning,papernot2016distillation,dziugaite2016study,hazan2016perturbations,guo2017countering,madry2017towards,ma2018characterizing,carlini2017towards,tramer2017ensemble,athalye2018obfuscated,uesato2018adversarial,vargas2019robustness}.

    \item[Zero-Shot learning:] 
    Zero-Shot learning is a method to estimate unfamiliar classes which do not appear in the training data.
    The motivation of Zero-Shot learning is to transfer knowledge from recognised classes to unfamiliar classes.
    Existing methods address the problem by estimating unfamiliar classes from an attribute vector defined manually for both known and unknown classes.
    % For each class, whether such an attribute (like colour, shape) relates to the class or not is represented by one or zero.
    The authors of \cite{lampert2009learning} introduced \emph{Direct Attribute Prediction (DAP)} model, which learns each parameter of the input sample for estimating the attributes of the sample from the feature vector generated.
    % It estimates an unknown class of the source data which is estimated from the target data by using these parameters. 
    % This approach projects generated feature vector into the source domain to classify the unknown classes.
    Based on this research, other zero-shot learning methods have been proposed which uses an embedded representation generated using a natural language processing algorithm instead of a manually created attribute vector \cite{norouzi2013zero,zhang2015zero,fu2015transductive,akata2015evaluation,zhang2016zero,bucher2016improving}.
    The opposite direction was proposed in \cite{shigeto2015ridge}, which learned how to project from the source domain to the generated feature vector.
    % In \cite{zhang2015zero}, a different strategy is proposed by constructing the histogram of known classes distribution for an unknown class to estimate unknown classes. 
    % They also assume that the unknown classes are the same if these histograms generated in the prediction domain and the source domain are similar.
    Here, we use a modified Zero-Shot learning approach entitled Raw Zero-Shot to understand the representation quality of the neural networks. 
    Our Raw Zero-Shot is distinguished from other zero-shot learning algorithms as in Raw Zero-Shot the neural network has no access to features (attribute vector) or additional supplementary knowledge. 
    \end{description}

\section{Raw Zero-Shot}
\label{section_raw_zero}

    \begin{wrapfigure}[15]{R}{0.65\linewidth}
        \centering
        \includegraphics[width=\linewidth]{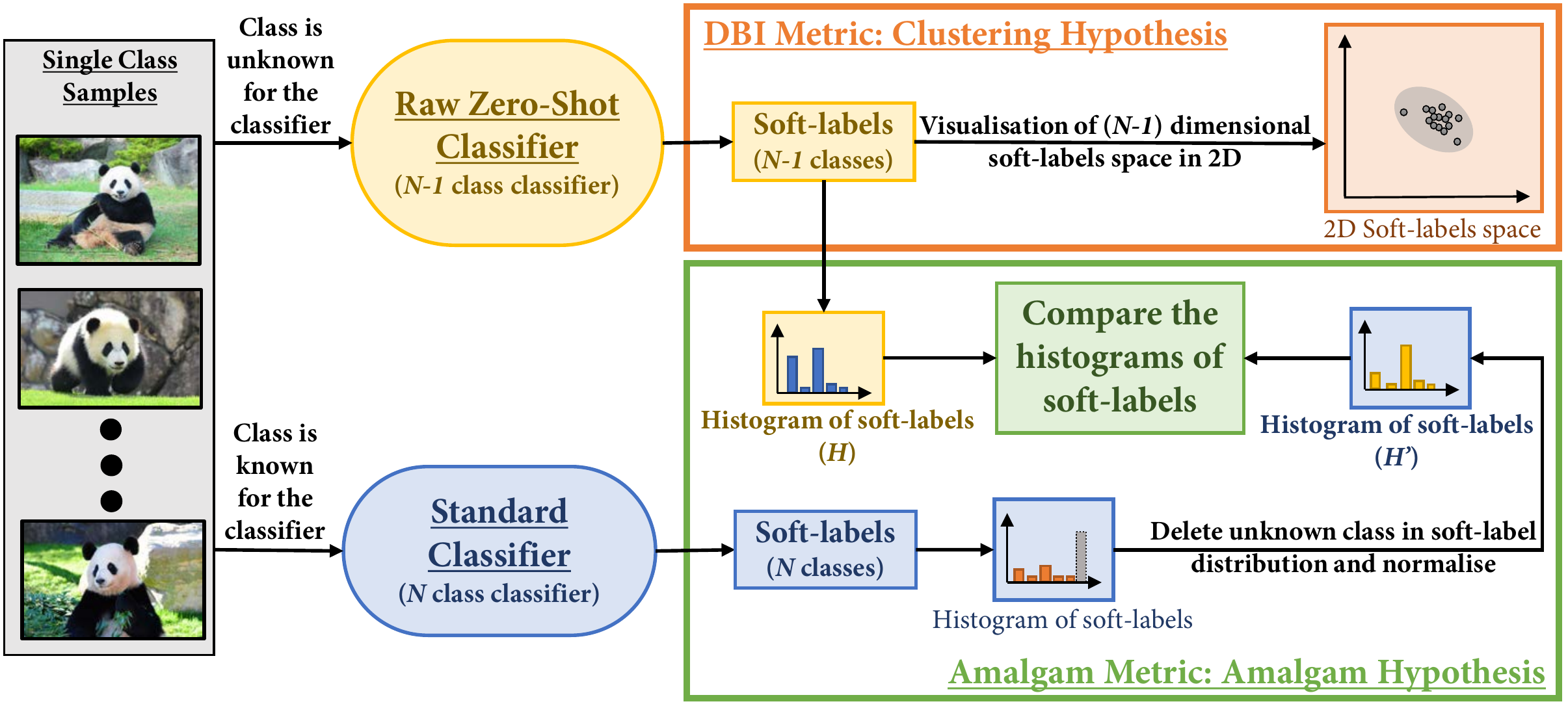} 
        \caption{Illustration of both Raw Zero-Shot Metrics.}
        \label{figure_illustration} 
    \end{wrapfigure}

    The Raw Zero-Shot is a supervised learning test in which only $N-1$ of the $N$ classes in the dataset are presented to the classifier during training.
    For training Raw Zero-Shot classifiers, all the samples of one specific class are removed from the standard training dataset.
    The Raw Zero-Shot classifier, thus, has only $N-1$ possible output in the form of soft-labels.
    Please note that a standard classifier has $N$ dimensions in the soft-label $(z)$ space. 
    In contrast, a Raw Zero-Shot classifier has $N-1$ dimensions in the soft-label $(z)$ space due to the exclusion of a class.
    During testing, only the unknown class (excluded class from $N$) is provided to the classifier.
    The output of the soft-labels for the given unknown class is recorded.
    This process is iterated for all potential $(N)$ classes, excluding a different class each time.

    Raw Zero-Shot Metrics are then computed over the soft-labels of the unknown (excluded) class to assess the representation quality (Figure \ref{figure_illustration}).
    These metrics are each based on a different hypothesis of what defines a feature or a class.
    In the same way, as there are various sorts of robustness, \textit{there are also different variations of representation quality}.
    Therefore, \textit{our metrics are complementary, each highlighting a different perspective of the whole}.
    The following subsections define them.

\subsection{Davies-Bouldin Metric (DBM) - Clustering Hypothesis}
\label{section_cluster_hypothesis}

    Soft labels of a classifier compose a space in which a given image would be categorised as a weighted vector involving the previously learned classes, similar to our example in Figure \ref{figure_overview}.
    In our example, the unknown class (Giant Panda) is represented as a combination of previously recognised classes (Bear, Zebra, Bird) where the feature (body shape) of the Bear is associated with the Giant Panda.
    Moreover, the feature (colour) of the Zebra is also associated with the Giant Panda. 
    This is analogous to how children describe previously unseen (Giant Panda) objects as a combination of previously seen objects (Bear and Zebra).
    Thus, all the images of the class Giant Panda should have similar soft-labels as the classifier can associate Giant Panda with some features of Zebra and Bear classes.

    Considering that the cluster of soft-labels of an unfamiliar class would constitute a class in itself, we can use cluster validation techniques to assess the representation.
    Here we choose for simplicity one of the most used metric in internal cluster validation, Davies-Bouldin Index \cite{davies1979cluster}.
    Hence, Davies-Bouldin Metric (DBM) for an unknown class is defined as follows:
    \begin{equation*} \text{DBM} = \left(\frac{1}{n} \sum_{j=1}^{n} {\left| z_j - G \right|^2}\right)^{1/2} \end{equation*}
    in which 
    $n$ is the number of samples (samples from unknown class), 
    $G$ is the centroid of the cluster formed by the soft-labels of all the $n$ samples, 
    $z$ is soft-label of a single sample of unknown class.
    A lower DBM Score, thus, corresponds to a better dense cluster formed by the soft-labels.

\subsection{Amalgam Metric (AM) - Amalgam Hypothesis}
\label{section_amalgam_hypothesis}

    If neural networks can learn the features existing in the classes, it would be reasonable to consider that the soft-labels also describe a given image as an aggregate of the previously learned classes (Figure \ref{figure_overview}). 
    We call this aggregate, Amalgam Proportion which is used in our metrics.
    Similar to a vector space in linear algebra, the soft-labels can be combined to describe unknown objects in this space.
    From our example, the unknown (Giant Panda) class can be represented as a combination of learned (Bear and Zebra) classes, where Giant Panda is associated with roughly $60\%$ of the features of the Bear class and $39\%$ of the features of the Zebra class.

    Differently from the previous metric, here we are interested in the exact values of the soft-labels to determine the best aggregate to encode an unknown class.
    However, what would constitute the true soft-label (Amalgam Proportion) for a given unfamiliar class still needs to be determined.
    To calculate the true soft-label of a given unknown class automatically, we use here the assumption that \textit{standard} classifiers (classifiers trained on the unknown class or in other words classifier trained on $N$ classes) should output a good approximation of the Amalgam Proportion. 
    This is based on the hypothesis that the features that are learned by a classifier share some similarity with the unfamiliar class and the classifier can associate this similarity in its feature space while evaluating these unfamiliar classes. 
    Therefore, if a standard classifier is trained in the $N$ classes, the soft-labels of the remaining $N-1$ classes (excluding the unknown class) is the validation Amalgam Proportion (Figure \ref{figure_illustration}).
    Consequently, the Amalgam Metric (AM) for an unknown class is defined as: 
    \begin{equation*} \text{AM} = \frac{\left\lVert H' - H \right \lVert_1}{N-1} \quad \text{where} \quad H = \sum_{j=1}^{n} z_j , \quad  H' = \sum_{j=1}^{n} z'_j \end{equation*}
    in which, $z'$ is the normalized soft-label from the standard classifier which is also the ground truth (true Amalgam Proportion) for the image of the unknown class, and $z$ is the soft-labels from the Raw Zero-Shot classifier which is also the estimated Amalgam Proportion for the unknown class.
    A lower AM Score, thus, corresponds to a better prediction of Amalgam Proportion which is closer to the ground truth. 

\section{On Link Between Representation Quality And Adversarial Attacks} 
\label{section_theory}
    
    We begin our theoretical analysis by developing a framework, loosely based on the settings proposed in \cite{madry2017towards,ilyas2019adversarial}.
    In the canonical multi-class classification setting, the goal of a classifier is to achieve low-expected loss: 
    \[ \standardloss{x}{y} \]  
    for the given input-label pair $(x, y) \in \mathcal{X} \times [\![1..N]\!] $ belonging to a distribution $\dist$ where $N$ is the number of classes whereas a classifier in the robustness evaluation setting should have a low-expected adversarial loss: 
    \[ \adversarialloss{x}{y} \] 
    where $\Delta$ represents an appropriately defined set of perturbations that an adversary can apply to induce misclassification\footnote{Here, we use the error in noise to be the adversarial loss instead of the worst-case error, for a discussion of the relationship between error in noise and adversarial samples, please refer to \cite{gilmer2019adversarial}.}.
    It is stated in \cite{ilyas2019adversarial} that the adversarial vulnerability is a significant consequence of the dominant supervised learning paradigm as we usually train classifiers to maximise (distributional) accuracy on $y$. 
    
    To circumvent this bias towards supervised learning, we take into account the representation-of-features $z$ learned by the classifier.
    Thus, the expected adversarial loss as defined over representation-of-features $z$ becomes:
    \[ \adversarialloss{x}{z} \] 
    The intuition to take into account the representation-of-features $z$ is to analyse a classifier beyond the paradigm of supervised learning and to analyse the association of this representation-of-features $z$ to an unfamiliar class.
    Note that as an ideal representation-of-features is not defined for an unfamiliar class; therefore, we assess $z$ using unsupervised learning evaluation. 

    Theoretically, any projection of the input in any of the feature space of the classifier could be used as $z$ to assess the representation quality of the classifier.
    Here, we use the final projection of the input as representation-of-features $z$, as final projection is usually a reasonable estimate of the entire projection learned by the classifier.
    % In $z$, it is also possible to evaluate the representation of known and unknown classes which should share some features.
    We use our Davies-Bouldin Metric (DBM) (Section \ref{section_cluster_hypothesis}) to evaluate cluster of an unfamiliar class in representation-of-features space $z$ by its intra-cluster distance.
    Further, we also evaluate the relationship between representation-of-features $z$ and an approximated ground truth of representation-of-features using our other Amalgam Metric (AM) (Section \ref{section_amalgam_hypothesis}).
    % Moreover, we hypothesise that unknown classes should evaluate $z$ with less bias because a direct map from input to output is inexistent, being orthogonal to the learning objective.

\section{Analysis of Representation Quality}

\subsection{Experimental Design}

    \begin{description}[leftmargin=*]

    \item[Considered Datasets.] 
    We conducted the experiments based on Raw Zero-Shot on three diverse datasets to evaluate the representation of the neural networks.
    We used Fashion MNIST (F-MNIST) \cite{xiao2017fashion}, standard CIFAR-10 dataset \cite{krizhevsky2009learning} and a customised Sub-Imagenet (Sub) dataset for our evaluations. 
    The details of the customised Sub-Imagenet dataset is mentioned in Appendix \ref{section_details_sub}.
    Note that, the number of samples ($7000$ for Fashion MNIST, $6000$ for CIFAR-10, and roughly $13500$ samples for Sub-Imagenet dataset) in the assumed unknown class differ with the dataset. 
    We use the samples from both training and testing dataset for the unknown class for experimentation because we exclude these samples in the training process.

    \item[Considered Classifiers.] 
    To obtain the results on vanilla classifiers, we evaluate different classifiers for different datasets. 
    For the Fashion MNIST datasets, we chose to evaluate two different architecture types of classifiers, 
    a Multi-Layer Perceptron (MLP), and 
    a shallow Convolution Neural Network (ConvNet).
    For the CIFAR-10 dataset, we chose to evaluate a wide range of classifiers such as 
    Capsule Networks (CapsNet) (a recently proposed completely different architecture based on dynamic routing and capsules) \cite{sabour2017dynamic}, 
    Residual Networks (ResNet) (a state-of-the-art architecture based on skip connections) \cite{he2016deep}, 
    Wide Residual Networks (WideResNet) (an architecture which also expands in width) \cite{zagoruyko2016wide},
    DenseNet (a state-of-the-art architecture which is a logical extension of ResNet) \cite{huang2017densely},
    Network in Network (NIN) (an architecture which uses micro neural networks instead of linear filters) \cite{lin2013network}, 
    All Convolutional Network (AllConv) (an architecture without max pooling and fully connected layers) \cite{springenberg2014striving}, 
    VGG-16 (a previous state-of-the-art architecture which is also a historical mark) \cite{simonyan2014very}, and 
    LeNet (a simpler architecture which is also a historical mark) \cite{lecun1998gradient}.
    For our Sub-Imagenet dataset, we chose two widely popular architectures, 
    InceptionV3 \cite{szegedy2016rethinking}, and 
    ResNet-50 \cite{he2016deep}.

    \item[Considered Adversarial Defences.]
    We also evaluate the representation quality of some of the adversarial defences for CIFAR-10, such as
    Feature Squeezing (FS) \cite{xu2017feature}, 
    Spatial Smoothing (SS) \cite{xu2017feature} ,
    Label Smoothing (LS) \cite{hazan2016perturbations}, and
    Thermometer Encoding (TE) \cite{buckman2018thermometer}.
    We also evaluate classifiers trained with augmented dataset having Gaussian Noise of $\sigma = 1.0$ (G Aug).

    \item[Considered Attacks.]
    We evaluated all our vanilla classifiers against well-known adversarial attacks such as \nocite{art2018}
    Fast Gradient Method (FGM) \cite{goodfellow2014explaining}, 
    Basic Iterative Method (BIM) \cite{kurakin2016adversarial}, 
    Projected Gradient Descent Method (PGD) \cite{madry2017towards}, 
    DeepFool \cite{moosavi2016deepfool}, and 
    NewtonFool \cite{jang2017objective}. 
    Details about the adversarial attacks used are mentioned in Appendix \ref{section_details_attack}.

    \end{description}

\subsection{Experimental Results And Analysis Of Representation Quality} 
\label{section_results}

    \begin{table*}[!]
        \centering
        \resizebox{\linewidth}{!}{
        \begin{tabular}{l|r|r|rrrr}
        \toprule
        \textbf{Classifier} & \multicolumn{1}{c|}{\textbf{No Defence}} & \multicolumn{1}{c|}{\textbf{G Aug}} & \multicolumn{1}{c}{\textbf{FS}} & \multicolumn{1}{c}{\textbf{SS}} & \multicolumn{1}{c}{\textbf{LS}} & \multicolumn{1}{c}{\textbf{TE}}\\
        \midrule
        \multicolumn{7}{c}{\textbf{DBM}} \\
        \midrule
        LeNet      & 0.54 $\pm$ 0.04 & 0.56 $\pm$ 0.04 & 0.48 $\pm$ 0.17 & 0.45 $\pm$ 0.15 & 0.39 $\pm$ 0.13 & 0.52 $\pm$ 0.04 \\
        AllConv    & 0.64 $\pm$ 0.08 & 0.64 $\pm$ 0.11 & 0.59 $\pm$ 0.21 & 0.56 $\pm$ 0.20 & 0.44 $\pm$ 0.15 & 0.67 $\pm$ 0.05 \\
        NIN        & 0.63 $\pm$ 0.09 & 0.64 $\pm$ 0.11 & 0.58 $\pm$ 0.21 & 0.58 $\pm$ 0.20 & 0.48 $\pm$ 0.17 & 0.65 $\pm$ 0.06 \\
        ResNet     & 0.64 $\pm$ 0.13 & 0.63 $\pm$ 0.14 & 0.59 $\pm$ 0.23 & 0.59 $\pm$ 0.22 & 0.50 $\pm$ 0.20 & 0.71 $\pm$ 0.06 \\
        DenseNet   & 0.61 $\pm$ 0.14 & 0.60 $\pm$ 0.15 & 0.56 $\pm$ 0.23 & 0.57 $\pm$ 0.22 & 0.51 $\pm$ 0.21 & 0.69 $\pm$ 0.09 \\
        WideResNet & 0.58 $\pm$ 0.15 & 0.52 $\pm$ 0.23 & 0.54 $\pm$ 0.23 & 0.55 $\pm$ 0.21 & 0.42 $\pm$ 0.17 & 0.66 $\pm$ 0.08 \\
        VGG-16     & 0.61 $\pm$ 0.12 & 0.63 $\pm$ 0.12 & 0.56 $\pm$ 0.22 & 0.56 $\pm$ 0.20 & 0.51 $\pm$ 0.19 & 0.65 $\pm$ 0.05 \\
        CapsNet    & 0.43 $\pm$ 0.03 & 0.44 $\pm$ 0.03 & 0.39 $\pm$ 0.13 & 0.38 $\pm$ 0.13 & 0.38 $\pm$ 0.03 & 0.35 $\pm$ 0.07 \\
        \midrule
        \multicolumn{7}{c}{\textbf{AM}} \\
        \midrule
        LeNet      & 473.97 $\pm$ 91.53 & 491.10 $\pm$  84.92 & 429.56 $\pm$ 137.53 & 404.50 $\pm$ 90.53  & 392.50 $\pm$ 101.83 & 400.11 $\pm$  95.89 \\
        AllConv    & 634.04 $\pm$ 22.01 & 639.39 $\pm$  26.71 & 571.77 $\pm$ 184.66 & 534.96 $\pm$ 149.44 & 521.57 $\pm$ 150.31 & 590.91 $\pm$  45.64 \\
        NIN        & 646.04 $\pm$ 16.40 & 651.94 $\pm$  10.92 & 581.09 $\pm$ 191.16 & 551.55 $\pm$ 165.03 & 552.39 $\pm$ 168.52 & 620.91 $\pm$  23.65 \\
        ResNet     & 654.90 $\pm$ 6.40  & 656.61 $\pm$   4.82 & 589.55 $\pm$ 194.20 & 505.77 $\pm$ 120.16 & 538.53 $\pm$ 154.68 & 634.74 $\pm$  12.43 \\
        DenseNet   & 658.21 $\pm$ 4.05  & 659.28 $\pm$   4.18 & 592.64 $\pm$ 195.52 & 523.19 $\pm$ 113.68 & 540.31 $\pm$ 157.50 & 644.77 $\pm$   8.49 \\
        WideResNet & 660.00 $\pm$ 3.60  & 594.10 $\pm$ 197.11 & 594.20 $\pm$ 196.55 & 505.82 $\pm$ 113.56 & 542.77 $\pm$ 158.33 & 648.67 $\pm$   7.80 \\
        VGG-16     & 645.86 $\pm$ 15.19 & 649.83 $\pm$  11.12 & 582.56 $\pm$ 189.08 & 533.88 $\pm$ 144.35 & 533.93 $\pm$ 153.55 & 608.25 $\pm$  30.03 \\
        CapsNet    & 386.02 $\pm$ 82.02 & 383.92 $\pm$  72.53 & 425.73 $\pm$ 156.14 & 418.17 $\pm$ 180.07 & 380.23 $\pm$ 119.27 & 857.40 $\pm$ 198.97 \\
        \bottomrule
        \end{tabular}
        }
        \caption{Mean and Standard Devidation of DBM and AM values for different classifiers with and without the adversarial defences on CIFAR-10.}
        \label{table_result}
    \end{table*}

    \begin{table*}[!]
        \centering
        \resizebox{0.7\linewidth}{!}{
        \begin{tabular}{l|rr|rr}
        \toprule
        \multirow{2}{*}{\textbf{Metrics}}& \multicolumn{2}{c}{\textbf{Fashion MNIST}} & \multicolumn{2}{c}{\textbf{Sub Imagenet}} \\
        & \multicolumn{1}{c}{MLP} & \multicolumn{1}{c|}{ConvNet} & \multicolumn{1}{c}{InceptionV3} & \multicolumn{1}{c}{ResNet-50} \\
        \midrule
        \textbf{DBM} & 0.51 $\pm$ 0.09    & 0.47 $\pm$ 0.10    & 0.56 $\pm$ 0.07     & 0.55 $\pm$ 0.15     \\
        \textbf{AM}  & 670.71 $\pm$ 81.79 & 683.55 $\pm$ 76.39 & 1335.65 $\pm$ 31.83 & 1311.97 $\pm$ 37.59 \\
        \bottomrule
        \end{tabular}
        }
        \caption{Mean and Standard Devidation of DBM and AM values for different classifiers on different datasets.}
        \label{table_other_result}
    \end{table*}

    Table \ref{table_result} shows the results of the Raw Zero-Shot metrics (DBM and AM) for vanilla classifiers and vanilla classifiers employed with a variety of adversarial defences for improving the robustness of vanilla classifiers for CIFAR-10.
    Table \ref{table_other_result} shows the results of the Raw Zero-Shot metrics (DBM and AM) for vanilla classifiers for other datasets (Fashion MNIST and Sub Imagenet).
    Note that, we use mean of all the metric values for $N$ classes of the dataset to be characteristic metric value for a classifier.
    To enable the visualisation of the DBM metric, we plot a projection of all the points in the decision space of unknown classes ($N-1$ dimensions) into two-dimensional space for the vanilla classifiers (Appendix \ref{section_visualise_dbm}).
    We can also similarly visualise the Amalgam metric in the form of histograms of soft-labels for the vanilla classifiers (Appendix \ref{section_visualise_am}). 

    \paragraph{Broad Overview:} According to our metrics (Tables \ref{table_result} and \ref{table_other_result}) we summarise our results, 

    \begin{description}[before={\setcounter{descriptcount}{0}}, font=\bfseries\stepcounter{descriptcount}\thedescriptcount)~, leftmargin=*]
    \item[For CIFAR-10 dataset.] 
    CapsNet possesses the best representation amongst all classifiers examined as it has the least (best) score in both of our metrics. 
    At the same time, LeNet is considered the second-best neural network as it has the second-least (second-best) score for both of our metrics.

    \item[For Sub-Imagenet dataset.] 
    Both the architectures (InceptionV3 and ResNet-50) are equally clustered and predict the Amalgam Proportion similarly.
    However, ResNet-50 has marginally better representation than the InceptionV3 as it has better scores for both of our metrics. 

    \item[For Fashion MNIST dataset.]
    Interestingly, both the architectures (MLP and ConvNet) have a similar quality of representation.
    While ConvNet seems marginally superior to the MLP in terms of clustering the unknown classes more tightly (suggested by DBM), MLP seems marginally superior to predict the Amalgam Proportion better than the ConvNet (suggested by AM). 

    \item[For Adversarial Defences:]
    Adversarial defences, in general, improve the representation quality of the neural networks, either by creating more dense cluster of the soft-labels (suggested by DBM), or by providing better prediction of Amalgam Proportion (suggested by AM), or both.
    Hence, they are linked with the representation quality, and the results suggest that the current adversarial defences improve the robustness of the neural network by improving the representation quality. 

    \end{description}

    On carefully observing the metric values (Table \ref{table_result}), we found that our assessment of representation quality also explains some of the propositions by other researchers, we highlight our key findings below,

    \begin{description}[style=sameline, leftmargin=*]

    \item[Why LeNet has better representation quality than other deeper networks?] 
    The fact, that LeNet achieves a higher representation quality than other deeper neural networks such as ResNet and AllConv may seem extremely unlikely.
    However, accurate classifiers can trade-off robustness for accuracy \cite{tsipras2018robustness,raghunathan2020understanding}.
    Our metrics suggests that this trade-off happens because the representation quality of the deeper neural networks has worsened.

    \item[Does a model with high capacity will have a better representation?] 
    Our results reveal that a deeper network which generally has a higher capacity \cite{madry2017towards} does not necessarily correspond to a better representation of input features. 
    As CapsNet and LeNet, which are much shallower than the other deeper networks, are shown to have superior representation quality than other deeper networks.

    \item[How does Label Smoothing improve the representation quality?] 
    It is suggested in \cite{muller2019does}, that Label Smoothing (LS) encourage the representations to group in tight, equally distant clusters.
    The raw metric values for LS not only suggests that classifiers employed with LS form a tighter cluster in soft-label space (suggested by DBM), but also the prediction of Amalgam Proportion is close to the ground-truth (suggested by AM) than their vanilla counterparts. 
    Thus, our results (suggested by DBM) corroborate the results that LS encourage the representations to group in tight clusters \cite{muller2019does}.

    \item[How does augmenting the dataset with Gaussian Noise affect the representation?]
    We also observe that Gaussian augmentation degrades the representation quality of all the classifiers. 
    This supports our intuition (Section \ref{section_theory}), as adding Gaussian noise to the images subdue the features of the image by blurring making the classifier harder to interpret these features. 
    
    \item[Does Thermometer Encodding breaks linearity in neural networks?] 
    Interestingly, the result of the Thermometer Encoding (TE) suggests that when classifiers (except CapsNet) are trained with TE, they tend to predict closer to the ground truth (suggested by AM). 
    However, the soft-labels form a sparser cluster (suggested by DBM) than the vanilla counterparts.
    This suggests that due to the sparsity of input feature space created by TE \cite{buckman2018thermometer}, the output representation also gets sparsified.
    This is expected behaviour since TE tries to break the linearity \cite{goodfellow2014explaining} of the classifier. 
    Our metrics suggest that TE breaks this linearity of the classifier to some extent by creating sparser output representations. 

    \item[Why CapsNet has better representation quality than other deeper networks?] 
    We believe Capsule Networks has the best representation amongst other neural networks, because of the dynamical nature (routing) of the CapsNet. 
    Further, the use of non-linear squashing function \cite{sabour2017dynamic} in the CapsNet suggest that for breaking linearity \cite{goodfellow2014explaining} in neural networks; we do not have to compromise on the representation quality. 

    \item[Can we identify adversarial defences which work on the principle of obfuscated gradients using representation quality?] 
    The adversarial defences (except LS) fail to improve the AM Score for the CapsNet while improving the DBM score.
    This is expected since the adversarial defences tend to make better representations, in general, than their vanilla counterparts by forming tighter clusters of the soft-labels (suggested by DBM) and therefore the DBM results of the CapsNet follow the norm. 
    However, our intuition is that these defences fail to modify the gradients, due to the dynamical nature of the CapsNet, as intended and end up predicting the Amalgam Proportion farther from the ground truth (suggested by AM).
    Hence, this result suggests that these adversarial defences which work on modifying gradients (obfuscating gradients \cite{athalye2018obfuscated}) fail to work as intended in the case of dynamical classifiers, such as CapsNet and hence, can be identified with our metrics. 

    \end{description}

\section{The Link Between Representation Quality And Adversarial Attacks} 
\label{section_link_attack}

    \begin{table*}[!]
        \centering
        \resizebox{\linewidth}{!}{%
        \begin{tabular}{l|rrrrr|rrrrr}
            \toprule
            \multirow{2}{*}{\textbf{Classifier}} & \multicolumn{5}{c}{\textbf{DBM with Mean $L_2$ Score}} & \multicolumn{5}{|c}{\textbf{AM with Mean $L_2$ Score}} \\
            & \multicolumn{1}{c}{\textbf{FGM}} & \multicolumn{1}{c}{\textbf{BIM}} & \multicolumn{1}{c}{\textbf{PGD}} & \multicolumn{1}{c}{\textbf{DeepFool}} & \multicolumn{1}{c}{\textbf{NewtonFool}} 
            & \multicolumn{1}{|c}{\textbf{FGM}} & \multicolumn{1}{c}{\textbf{BIM}} & \multicolumn{1}{c}{\textbf{PGD}} & \multicolumn{1}{c}{\textbf{DeepFool}} & \multicolumn{1}{c}{\textbf{NewtonFool}} 
            \\
            \midrule
            \multicolumn{11}{c}{\textbf{Fashion MNIST}} \\
            \midrule 
            MLP           
                        & -0.20 (0.58) & -0.17 (0.64) & -0.17 (0.64) & -0.04 (0.91) & -0.02 (0.97) 
                        & 0.82 (0.00) & 0.26 (0.47) & 0.26 (0.47) & 0.83 (0.00) & 0.84 (0.00) 
                        \\
            ConvNet 
                        & -0.24 (0.50) & -0.30 (0.40) & -0.30 (0.40) & -0.26 (0.46) & -0.22 (0.55)
                        & 0.83 (0.00) & -0.07 (0.84) & -0.09 (0.80) & 0.81 (0.00) & 0.82 (0.00)  
                        \\
            \midrule
            \multicolumn{11}{c}{\textbf{CIFAR-10}} \\
            \midrule 
            LeNet      
                        & -0.18 (0.61) & -0.70 (0.02) & -0.66 (0.04) & -0.51 (0.13) & -0.36 (0.31) 
                        & 0.93 (0.00) & 0.32 (0.36)  & 0.25 (0.49)  & 0.81 (0.00)  & 0.89 (0.00)  
                        \\
            AllConv    
                        & -0.31 (0.39) & -0.56 (0.09) & -0.54 (0.11) & -0.10 (0.78) & -0.30 (0.41) 
                        & 0.67 (0.03) & 0.42 (0.23)  & 0.41 (0.24)  & 0.94 (0.00)  & 0.73 (0.02) 
                        \\
            NIN        
                        & -0.56 (0.09) & -0.57 (0.08) & -0.57 (0.09) & -0.42 (0.22) & -0.43 (0.21) 
                        & 0.78 (0.01) & 0.84 (0.00)  & 0.84 (0.00)  & 0.96 (0.00)  & 0.89 (0.00) 
                        \\
            ResNet     
                        & -0.52 (0.12) & -0.76 (0.01) & -0.76 (0.01) & -0.47 (0.17) & -0.51 (0.13) 
                        & 0.35 (0.32) & 0.57 (0.09)  & 0.57 (0.09)  & 0.79 (0.01)  & 0.83 (0.00)
                        \\
            DenseNet   
                        & -0.62 (0.06) & -0.50 (0.14) & -0.49 (0.15) & -0.16 (0.65) & -0.22 (0.55)
                        & 0.53 (0.11) & 0.78 (0.01)  & 0.78 (0.01)  & 0.78 (0.01)  & 0.84 (0.00)  
                        \\
            WideResNet 
                        & -0.68 (0.03) & -0.75 (0.01) & -0.75 (0.01) & -0.68 (0.03) & -0.75 (0.01) 
                        & 0.66 (0.04) & 0.68 (0.03)  & 0.68 (0.03)  & 0.78 (0.01)  & 0.68 (0.03) 
                        \\
            VGG-16     
                        & -0.62 (0.06) & -0.21 (0.55) & -0.20 (0.58) & -0.52 (0.13) & -0.63 (0.05) 
                        & 0.71 (0.02) & -0.04 (0.91) & -0.07 (0.85) & 0.87 (0.00)  & 0.74 (0.01) 
                        \\
            CapsNet    
                        & -0.71 (0.02) & -0.45 (0.19) & -0.49 (0.15) & -0.39 (0.26) & -0.48 (0.17) 
                        & 0.98 (0.00) & 0.69 (0.03)  & 0.73 (0.02)  & -0.17 (0.63) & 0.47 (0.17) 
                        \\
            \midrule 
            \multicolumn{11}{c}{\textbf{Sub-Imagenet}} \\
            \midrule
            InceptionV3 
                        & -0.76 (0.01) & -0.52 (0.13) & -0.52 (0.13) & -0.35 (0.32) & -0.50 (0.14) 
                        & 0.75 (0.01) & 0.14 (0.70) & 0.14 (0.70) & 0.28 (0.44) & 0.25 (0.49) 
                        \\
            ResNet-50   
                        & -0.34 (0.34) & -0.12 (0.74) & -0.12 (0.74) & -0.54 (0.10) & -0.25 (0.48) 
                        & 0.82 (0.00) & 0.31 (0.39) & 0.31 (0.39) & 0.51 (0.13) & 0.50 (0.15) 
                        \\
            \bottomrule
        \end{tabular}
        }
        \caption{Pearson correlation value (and p-value) of DBM and AM with Mean $L_2$ Score of Adversarial Attacks for each vanilla classifier and attack pair.}
        \label{table_pearson}
    \end{table*}

    Since, the results in Table \ref{table_result}, suggests a link between the representation quality and the adversarial defences as discussed above. 
    It is intuitive to assume that there exists a link between the representation quality and the adversarial attacks.
    To evaluate the statistical relevance of this link between representation quality and adversarial attacks, we conducted a Pearson Correlation test of Raw Zero-Shot metrics (DBM and AM) of the vanilla classifiers with adversarial attacks. 
    We use the analysis of adversarial attacks in the form of Mean $L_2$ Score ($L_2$ difference between the original sample and the adversarial one) to compute the correlation. 
    The Pearson correlations of the Raw Zero-Shot metrics (DBM and AM) with Mean $L_2$ Score is shown in Table \ref{table_pearson} for every architecture and attacks. 
    Moreover, these Pearson relationships between the Raw Zero-Shot metrics and Mean $L_2$ Score can also be visualised (Appendix \ref{section_visualise_pearson}). 
    We also analyse the impact of adversarial attacks on the correct class soft-label (Appendix \ref{section_cdiff}). 

    The purpose of evaluating the Mean $L_2$ Score as an adversarial metric is because the score effectively assesses the impact of the adversarial attacks on classifiers.
    Note that, some of the adversarial attacks may successfully perturb the same number of samples. 
    Therefore, the effectiveness of an attack cannot be justified with only adversarial accuracy.
    In our experiments, this is crucial as we need to analyse the effect of the adversarial algorithm on each of the class of the dataset.
    Also, some previous researches show that robustness differs across different classes of a classifier \cite{papernot2016limitations,pan2019identifying}.
    Therefore, here Mean $L_2$ score not only determines the alteration in the adversarial sample, but it also analyses the effectiveness of an attack across different classes.

    The correlational analysis of our metrics suggests a strong relationship between our Raw Zero-Shot metrics and the adversarial attacks in general.
    We do observe some anomalies in the Pearson correlation with AM of BIM and PGD attacks for the ConvNet, and VGG-16 network and DeepFool for CapsNet.
    These anomalies are studied in detail (Appendix \ref{section_visualise_pearson}) to understand their existence.
    Our analysis suggests that these anomalies exist due to abnormal behaviour of some classes.  
    On careful study, we note that for all the classes, BIM and PGD have similar Mean $L_2$ Score. 
    At the same time, the representation quality differs for some of the classes of these classifiers.
    While for the CapsNet, the Airplane class had very low Mean $L_2$ score suggesting less perturbation to misclassify despite a relatively close prediction of Amalgam Proportion to the ground-truth which was abnormal compared to the other classes in the same setting.
    These anomalies further suggest that ground truth Amalgam Proportion for some of the classes may not be inherently robust. 
    However, the study of these representation qualities of individual classes is beyond the scope of this article and hence, left as future work.

    On the other hand, DBM was shown to be related quite reasonably with the adversarial attacks. 
    This may be because it was shown that forcing a loss function to make features near to the feature centroid is beneficial against adversarial attacks in \cite{agarwal2019improving}.
    Our metric DBM, as it calculates precisely the closeness of the feature to the feature centroid, support the results in \cite{agarwal2019improving}.
    
\section{Conclusions}

    In this article, we propose, a novel zero-shot learning-based method, entitled Raw Zero-Shot, to assess the representation of the neural network classifiers.
    In order to assess the representation, two associated metrics are formally defined based on different hypotheses of representation quality.
    The results suggest that CapsNet (dynamic routing network) has the best representation quality amongst classifiers which calls for a more in-depth investigation of Capsule Networks.
    Also, results reveal that classifiers employed with adversarial defences have better representation in general than their vanilla counterparts suggesting that to improve the robustness of the classifiers, we have to improve the representation quality of the classifiers too.  
    Further, the strong correlation between the metrics and the adversarial attacks suggest a firm link between the representation quality and adversarial attacks.
    Results, thus, recommend that evaluation of the representation from both metrics (DBM and AM) are linked with the adversarial algorithms.
    % Moreover, the behaviour of different architectures spotted in the DBM in the Isomap plots shows that the metrics indeed capture the existing representation differences.

    Interestingly, our assessment of representation quality also helps to understand some of the investigations by other researchers in terms of representation quality stated below,
    \begin{enumerate}[leftmargin=*, itemsep=0pt, topsep=0pt, partopsep=0pt]
    \item The trade-off between robustness for accuracy \cite{tsipras2018robustness,raghunathan2020understanding} can be described with the help of representation quality. 
    \item High Capacity Classifiers \cite{madry2017towards} does not necessarily have good representation.
    \item Label Smoothing encourages the classifiers to group in tighter clusters \cite{muller2019does} (better representation) and hence, contribute towards robustness and accuracy.
    \item Thermometer Encoding breaks the linearity \cite{buckman2018thermometer} of the neural networks to some extent by forcing the classifier to create sparser output representations.
    \item Adversarial Defences which are based on the principle of obfuscating gradients \cite{athalye2018obfuscated} do not work as intended for the dynamical models.
    \item Forcing the features to be closer to the feature centroid \cite{agarwal2019improving} helps in increasing robustness of the neural networks.
    \end{enumerate}

    Hence, the proposed Raw Zero-Shot was able to assess and understand the representation quality of state-of-the-art neural networks, along with the adversarial defences and link the representation quality of the neural networks with adversarial attacks and defences.
    It also opens up new possibilities of using representation quality for both the evaluation (i.e. as a quality assessment) and the development (e.g. as a loss function) of neural networks.
    
    % % Furthermore, poor representation quality of deeper networks such as variants of ResNet suggests the need for non-linearity \cite{goodfellow2014explaining} and more dynamical classifiers \cite{vargas2019understanding} 
    % % (CapsNet, which is based on dynamic routing, has better representation quality than the other neural architectures).

% \setlength\intextsep{\oldintextsep}
% \setlength\textfloatsep{\oldtextfloatsep}

% \newpage
% \section*{Broader Impact}

% We believe this work will benefit the deep learning community in understanding the adversarial algorithms from the perspective of representation quality.
% We hope our empirical results can guide the community to develop better loss functions, optimisers, and other tools to improve the robustness of the neural networks.  
% Authors are required to include a statement of the broader impact of their work, including its ethical aspects and future societal consequences. 
% Authors should discuss both positive and negative outcomes, if any. For instance, authors should discuss 
% a) who may benefit from this research, 
% b) who may be put at disadvantage from this research, 
% c) what are the consequences of failure of the system, and 
% d) whether the task/method leverages biases in the data. 
% If authors believe this is not applicable to them, authors can simply state this.
% Use unnumbered first level headings for this section, which should go at the end of the paper. {\bf Note that this section does not count towards the eight pages of content that are allowed.}

\begin{ack}
This work was supported by JST, ACT-I Grant Number JP-50243 and JSPS KAKENHI Grant Number JP20241216.
Additionally, we would like to thank Prof. Junichi Murata for the kind support without which it would not be possible to conduct this research.
\end{ack}

\bibliographystyle{IEEEtrans}
{\small
\bibliography{../adversarial_machine_learning}
}

\newpage
\appendix

\section{Details About Customised Sub-Imagenet Dataset}
\label{section_details_sub}

\begin{table*}[!htb]
    \centering
    \resizebox{\linewidth}{!}{
    \begin{tabular}{l|rr|l}
    \toprule
    \textbf{Super-Classes} & \textbf{Training Images} & \textbf{Testing Images} & \textbf{Corresponding Imagenet (ILSVRC 2012) Classes} \\
    \midrule
    Automobile & 12981 & 500 & 407, 468, 555, 627, 654, 779, 817, 802, 866, 867 \\ 
    Ball       & 12971 & 500 & 429, 430, 522, 574, 722, 746, 768, 805, 852, 890 \\ 
    Bird       & 12990 & 500 & 7, 8, 9, 16, 22, 23, 24, 84, 94, 100             \\ 
    Dog        & 12904 & 500 & 205, 206, 207, 208, 209, 210, 211, 212, 213, 214 \\ 
    Feline     & 13000 & 500 & 283, 284, 285, 286, 287, 288, 289, 290, 291, 292 \\ 
    Fruit      & 12986 & 500 & 948, 949, 950, 951, 952, 953, 954, 955, 956, 957 \\ 
    Insect     & 12985 & 500 & 300, 301, 302, 303, 304, 305, 306, 307, 308, 309 \\ 
    Snake      & 12758 & 500 & 55, 56, 57, 58, 59, 60, 61, 62, 63, 64           \\
    Primate    & 12979 & 500 & 365, 366, 367, 368, 369, 370, 371, 372, 373, 374 \\ 
    Vegetable  & 12815 & 500 & 935, 936, 937, 938, 939, 943, 944, 945, 946, 947 \\ 
    \midrule
    \textbf{\textit{Total}} & \textbf{\textit{129359}} & \textbf{\textit{5000}} &  \\
    \bottomrule
    \end{tabular}
    }
    \caption{Description of Super-Classes used in the Sub-ImageNet.}
    \label{table_sub}
\end{table*}

Sub-Imagenet is subset of the Imagenet (ILSVRC 2012) \cite{russakovsky2015imagenet} dataset. 
It is intuitive for us to expect that as the number of classes $(N)$ grows, the decision boundary will become more complicated, causing the classifier to smoothen the representation (Amalgam Proportion) more.
Therefore, to prevent this bias, we grouped a subset of $100$ existing semantically alike ImageNet classes into $10$ distinct super-classes, as described in Table \ref{table_sub}.
Our Sub-Imagenet dataset has some desired characteristics for our experiments which are also similar to the CIFAR-10 dataset. 
These features are:

\begin{enumerate}[leftmargin=*, itemsep=0pt, topsep=0pt, partopsep=0pt]
\item It is relatively balanced dataset as other datasets used in the experiments. 
The dataset has a mean of $12937$ training images with a standard deviation of $80$ images. 
All super-classes have relatively the same number of images with a minimum of $12758$ images for super-class Snake and a maximum of $13000$ for super-class Feline.
Thus, the samples in the unknown class in our experiments remain relative same.
\item Type of super-classes is similar to CIFAR-10, having six animal classes and four non-animal classes. 
\item Abstract Relationships between super-classes also exists similar to the CIFAR-10.
The CIFAR-10 have a Cat-Dog and Automobile-Truck relationships in which they are semantically similar. 
Similarly, our Sub-imagenet also exhibits Dog-Feline and Fruit-Vegetable relationships. 
These abstract relationships are important to validate our hypothesis of Amalgam Proportion. 
\end{enumerate}

\section{Details About Adversarial Algorithms}
\label{section_details_attack}

\subsection{Experimental Settings}

\begin{table*}[!htb]
    \centering
    \resizebox{0.7\linewidth}{!}{
    \begin{tabular}{l|ll}
    \toprule
    \multirow{2}{*}{\textbf{Attack}} & \multirow{2}{*}{\textbf{For Fashion MNIST}} & \textbf{For CIFAR-10 and} \\
                                     &                                             & \textbf{Sub Imagenet}      \\
    \midrule
    FGM                           & $\text{norm} = L_\infty$, $\epsilon=0.3$, $\epsilon_{\text{step}}=0.01$  & $\text{norm} = L_\infty$, $\epsilon=8$, $\epsilon_{\text{step}}=2$  \\  
    \multirow{2}{*}{BIM}          & $\text{norm} = L_\infty$, $\epsilon=0.3$, $\epsilon_{\text{step}}=0.01$, & $\text{norm} = L_\infty$, $\epsilon=8$, $\epsilon_{\text{step}}=2$, \\ 
                                  & $\text{iterations} = 80$                                                 & $\text{iterations} = 10$ \\      
    \multirow{2}{*}{PGD}          & $\text{norm} = L_\infty$, $\epsilon=0.3$, $\epsilon_{\text{step}}=0.01$, & $\text{norm} = L_\infty$, $\epsilon=8$, $\epsilon_{\text{step}}=2$, \\ 
                                  & $\text{iterations} = 40$                                                 & $\text{iterations} = 20$                                            \\            
    DeepFool                      & $\text{iterations} = 100$, $\epsilon=0.02$                               & $\text{iterations} = 100$, $\epsilon=0.000001$                                 \\       
    NewtonFool                    & $\text{iterations} = 100$, $\text{eta}=0.375$                            & $\text{iterations} = 100$, $\text{eta}=0.01$                                            \\      
    \bottomrule
    \end{tabular}
    }
    \caption{Description of Adversarial Attack Parameters}
    \label{table_attack_params}
\end{table*}

All the adversarial attacks used in the article have been evaluated using Adversarial Robustness 360 Toolbox (ART v1.2.0) \cite{art2018}. 
We evaluated the test samples of Fashion MNIST, CIFAR-10 and Sub Imagenet datasets for the adversarial attacks.
The attack parameters used for the evaluated adversarial attacks are described in Table \ref{table_attack_params}.
Please note, that the adversarial settings were set such that the $L_2$ score of the adversarial attack was preferred over adversarial accuracy. 

\subsection{Results}

\begin{table*}[!htb]
    \centering
    \resizebox{0.95\linewidth}{!}{%
    \begin{tabular}{l|rrrrr|rrrrr}
        \toprule
        \multirow{2}{*}{\textbf{Classifier}} & \multicolumn{5}{|c}{\textbf{Adversarial Accuracy}} & \multicolumn{5}{|c}{\textbf{Mean $L_2$ Score}} \\
        & \multicolumn{1}{|c}{\textbf{FGM}} & \multicolumn{1}{c}{\textbf{BIM}} & \multicolumn{1}{c}{\textbf{PGD}} & \multicolumn{1}{c}{\textbf{DeepFool}} & \multicolumn{1}{c}{\textbf{NewtonFool}} 
        & \multicolumn{1}{|c}{\textbf{FGM}} & \multicolumn{1}{c}{\textbf{BIM}} & \multicolumn{1}{c}{\textbf{PGD}} & \multicolumn{1}{c}{\textbf{DeepFool}} & \multicolumn{1}{c}{\textbf{NewtonFool}} 
        \\
        \midrule
        \multicolumn{11}{c}{\textbf{Fashion MNIST}} \\
        \midrule 
        MLP     & 91.08 & 91.29 & 91.29 & 27.16    & 25.39  & 210.73 & 638.83 & 638.83 & 309.41   & 289.28     \\          
        ConvNet & 86.89 & 89.20 & 89.18 & 23.63    & 22.67  & 306.25 & 669.56 & 665.76 & 314.81   & 263.65     \\    
        \midrule
        \multicolumn{11}{c}{\textbf{CIFAR-10}} \\
        \midrule 
        LeNet      & 84.58 & 89.12 & 89.25 & 31.70    & 84.12      & 152.37 & 345.27 & 357.34 & 132.32   & 49.61     \\
        AllConv    & 67.09 & 69.11 & 69.11 & 51.46    & 61.86      & 155.95 & 273.90 & 274.15 & 487.46   & 61.05     \\
        NIN        & 72.49 & 74.26 & 74.26 & 59.94    & 66.76      & 140.46 & 216.97 & 216.96 & 492.90   & 54.78     \\
        ResNet     & 52.75 & 55.41 & 55.41 & 58.71    & 54.39      & 124.70 & 164.64 & 164.64 & 458.57   & 51.56     \\
        DenseNet   & 50.78 & 52.11 & 52.11 & 60.83    & 50.81      & 120.03 & 160.34 & 160.38 & 478.03   & 53.89     \\
        WideResNet & 69.59 & 89.42 & 89.44 & 60.10    & 82.73      & 159.88 & 208.44 & 208.49 & 613.14   & 63.13     \\
        VGG-16     & 82.79 & 94.97 & 94.99 & 65.08    & 92.43      & 181.29 & 321.86 & 329.96 & 651.65   & 77.01     \\
        CapsNet    & 70.02 & 82.23 & 84.46 & 87.40    & 90.04      & 208.89 & 361.63 & 370.90 & 258.08   & 1680.83   \\
        \midrule 
        \multicolumn{11}{c}{\textbf{Sub-Imagenet}} \\
        \midrule
        InceptionV3 & 85.76 & 87.24 & 87.24 & 86.94    & 58.44     & 796.53 & 1204.01 & 1204.01 & 609.54   & 319.73     \\
        ResNet-50   & 85.74 & 86.72 & 86.72 & 84.78    & 60.84     & 826.06 & 1264.30 & 1264.34 & 633.30   & 336.80     \\
        \bottomrule
    \end{tabular}
    }
    \caption{Adversarial Accuracy and Mean $L_2$ Score for each classifier and adversarial attack pair.}
    \label{table_attack}
\end{table*}

Table \ref{table_attack} shows the Adversarial Accuracy, and Mean $L_2$ Score, of different adversarial attacks for different classifiers.
Note that, the adversarial accuracy of some attacks (Table \ref{table_attack}) differ from the original published adversarial accuracy due to the change of parameters of the attacks mentioned in Table \ref{table_attack_params}. 
% A reason why we think LeNet might be easier to attack (or less robust), despite having a good representation maybe, because the search space is less complicated, which corresponds to less obfuscation \cite{athalye2018obfuscated}) and more limited capacity \cite{madry2017towards}. 
% In this case, the lack of robustness of LeNet is not because of its poor representation quality but because it is easier to find adversarial samples in shallower architectures (less complicated search space).

\section{Visualisations Of Davies-Bouldin Metric (DBM)}
\label{section_visualise_dbm}

\begin{figure*}[!htb]
    \centering
    \includegraphics[width=0.15\linewidth]{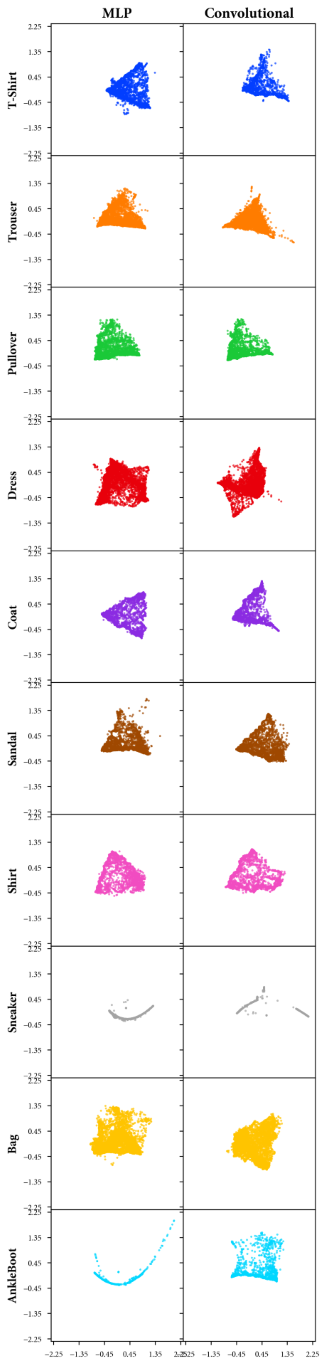} 
    \includegraphics[width=0.52\linewidth]{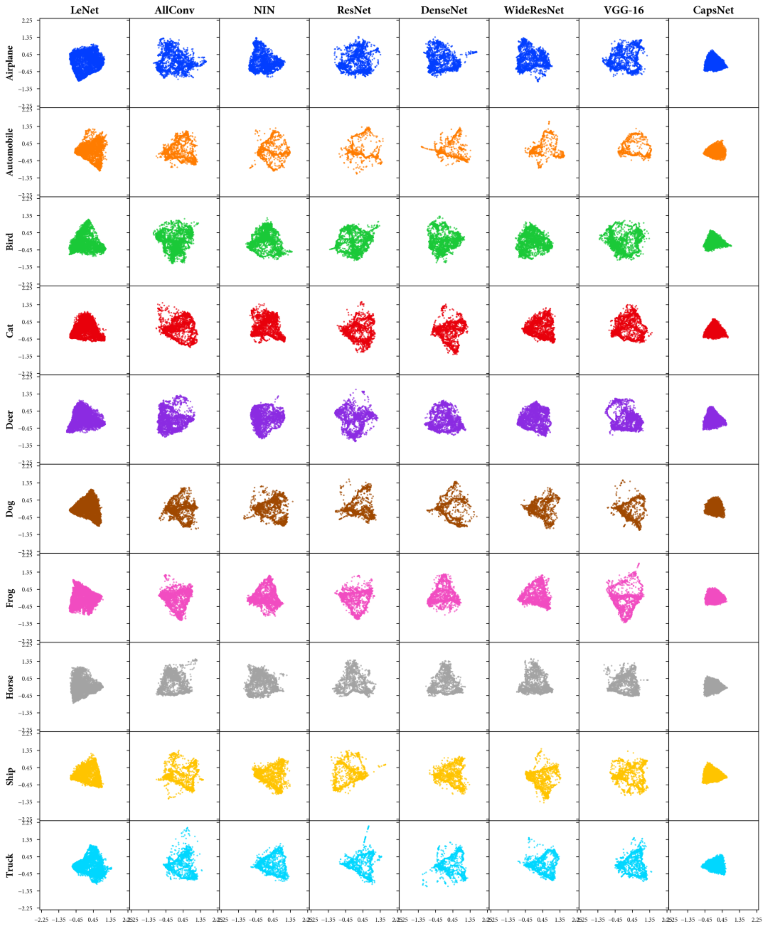}
    \includegraphics[width=0.15\linewidth]{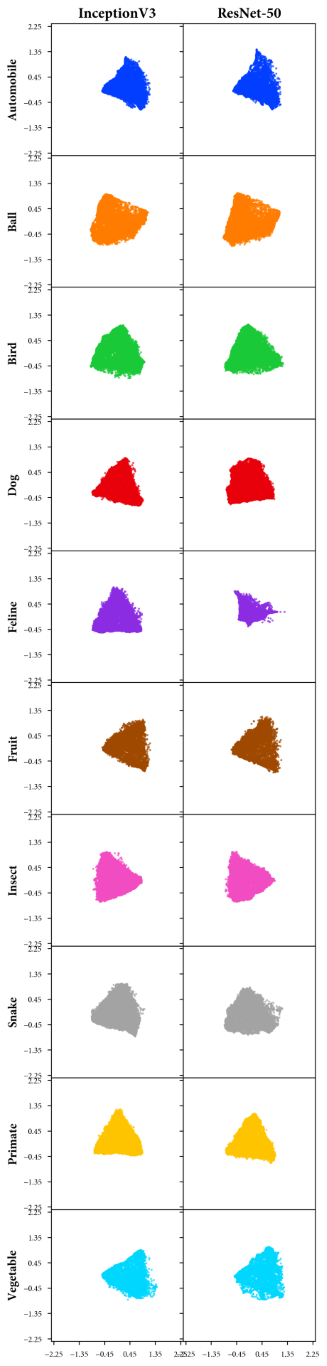}         
    \caption{Visualisation of the DBM results for vanilla classifiers using a topology preserving two-dimensional projection with Isometic Mapping (IsoMap). Each row represents a classifier trained with a label excluded whose projection is visualised.
    }
    \label{figure_isomap} 
\end{figure*}

\begin{figure*}[!htb]
    \centering
    \includegraphics[width=0.17\linewidth]{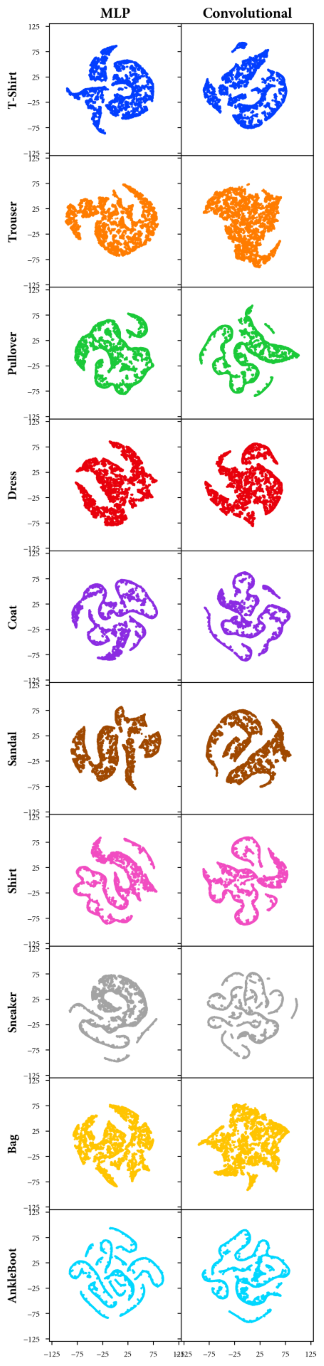} 
    \includegraphics[width=0.592\linewidth]{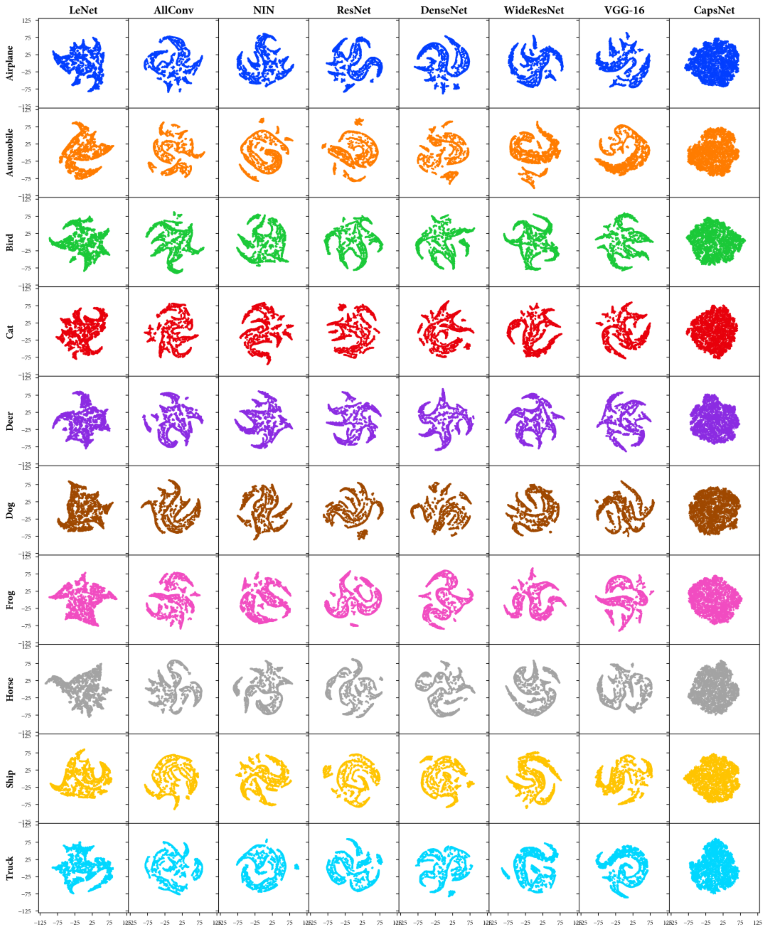}
    \includegraphics[width=0.17\linewidth]{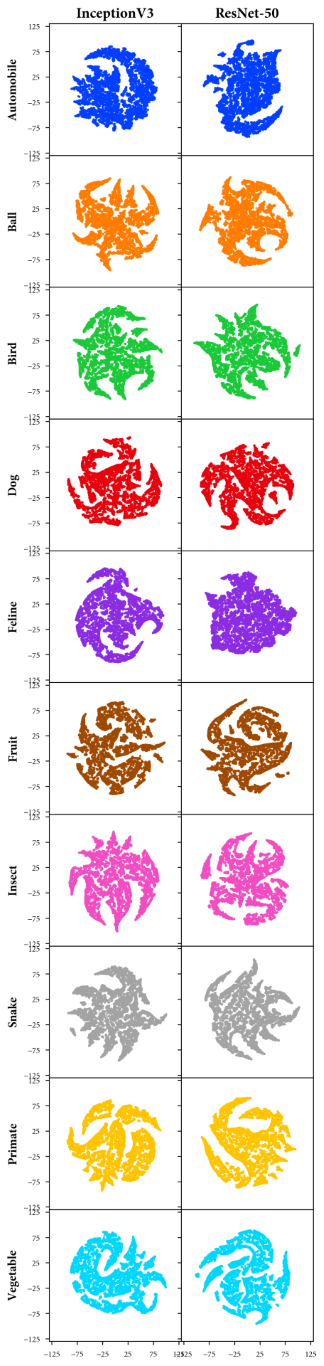} 
    \caption{Visualisation of the DBM results for vanilla classifiers using t-Distributed Stochastic Neighbour Embedding (t-SNE). Each row represents a classifier trained with a label excluded whose projection is visualised.}
    \label{figure_tsne}  
\end{figure*}

\begin{figure*}[!htb]
    \centering
    \includegraphics[width=0.17\linewidth]{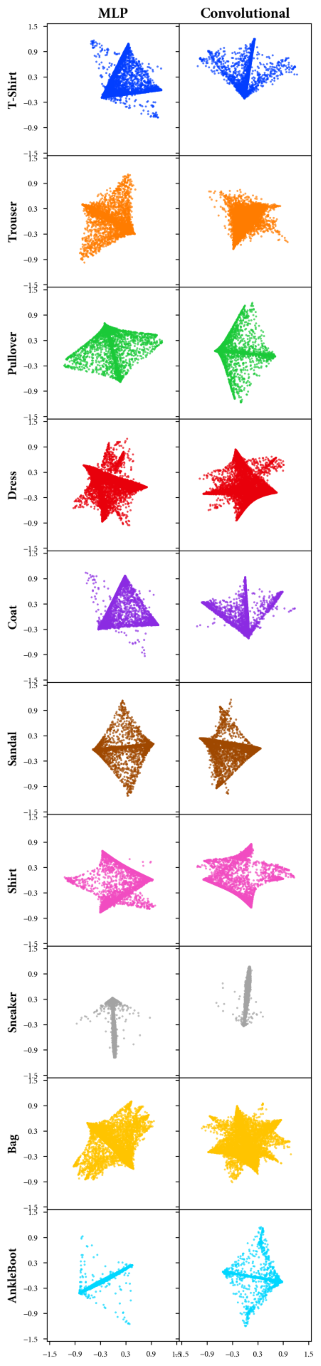} 
    \includegraphics[width=0.595\linewidth]{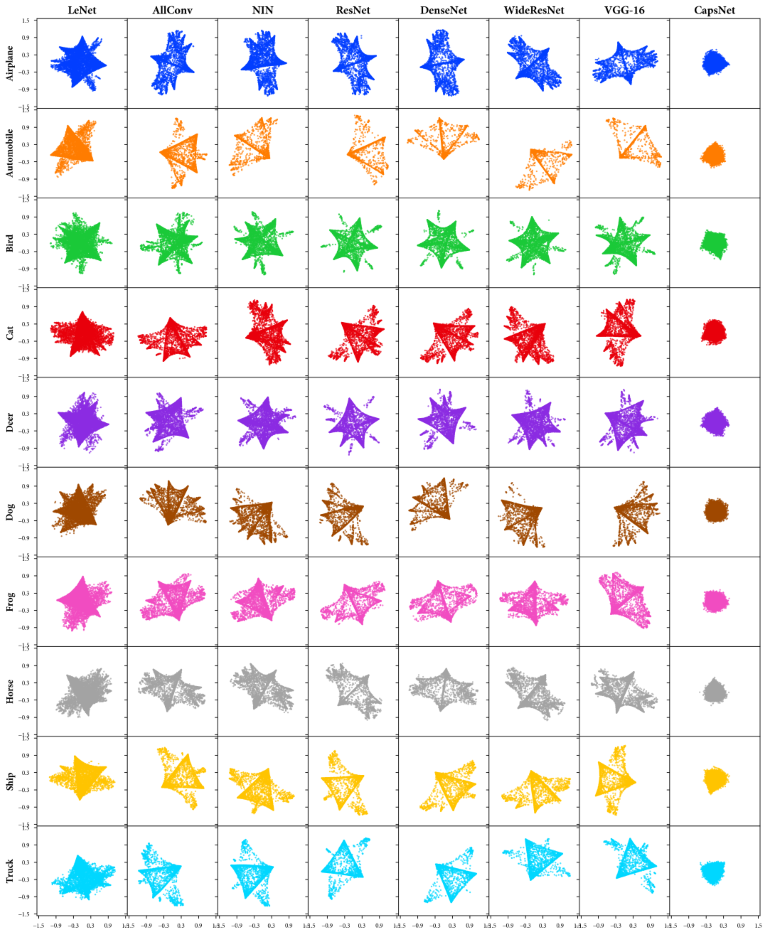}
    \includegraphics[width=0.17\linewidth]{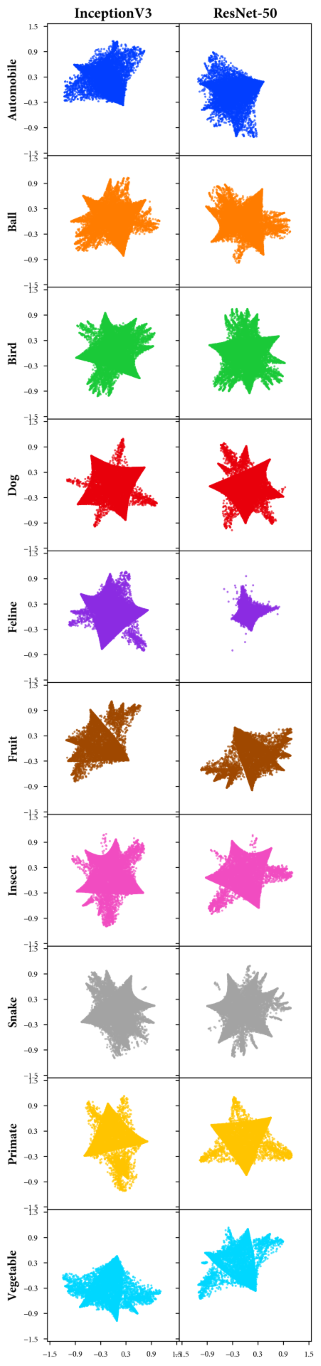} 
    \caption{Visualisation of the DBM results for vanilla classifiers using Multi Dimensional Scaling (MDS). Each row represents a classifier trained with a label excluded whose projection is visualised.}
    \label{figure_mds}   
\end{figure*}

\begin{figure*}[!htb]
    \centering
    \includegraphics[width=0.17\linewidth]{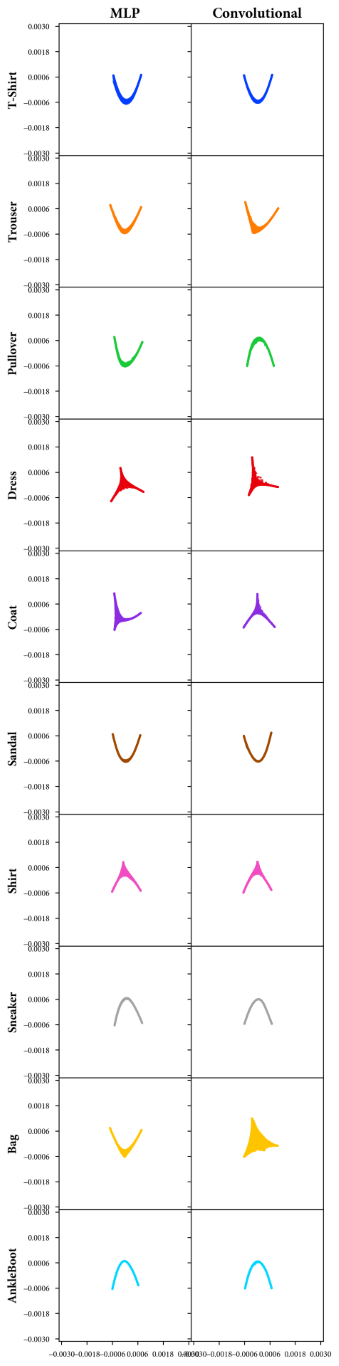} 
    \includegraphics[width=0.575\linewidth]{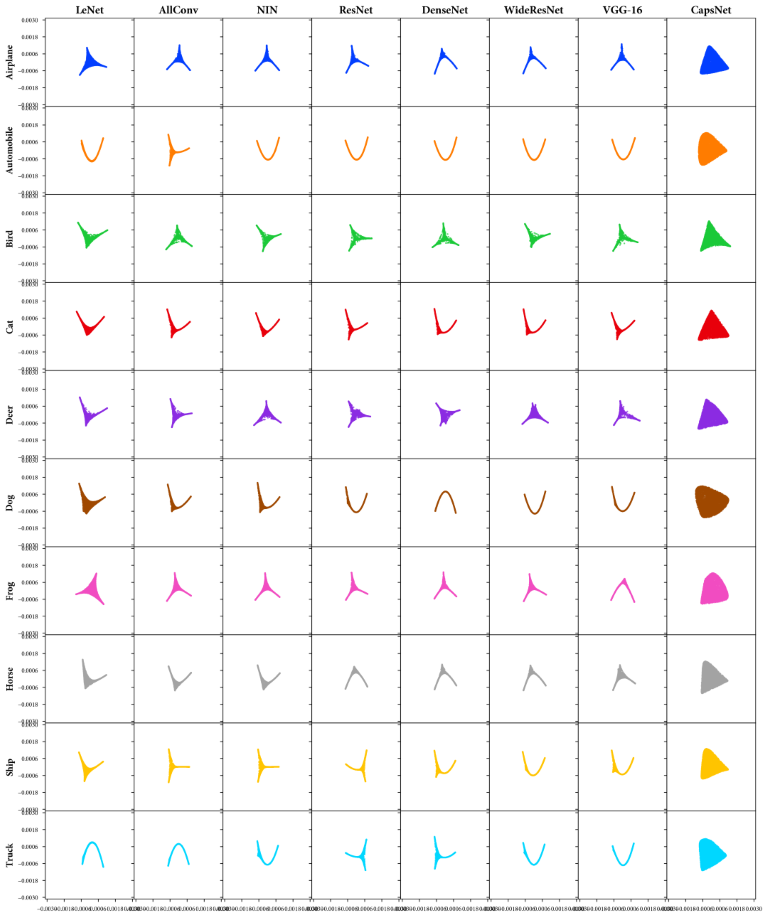}
    \includegraphics[width=0.17\linewidth]{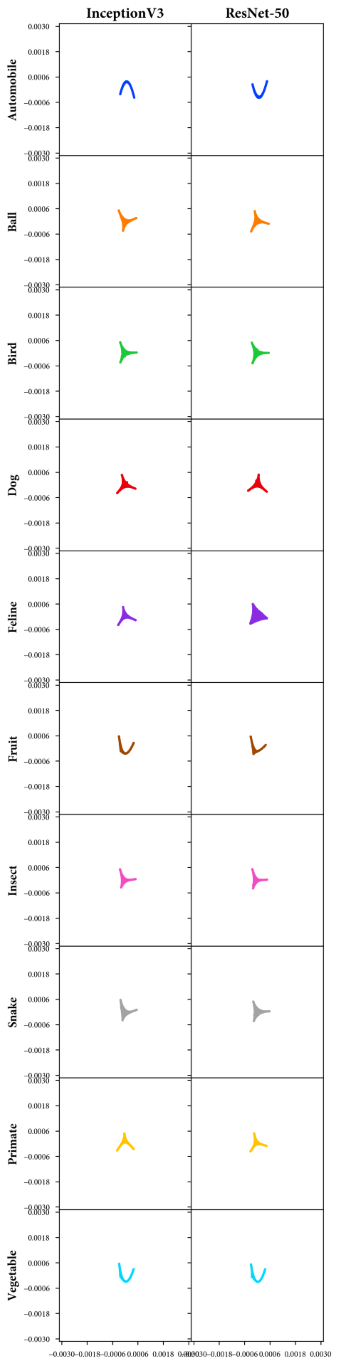} 
    \caption{Visualisation of the DBM results for vanilla classifiers using Spectral Embedding (SE). Each row represents a classifier trained with a label excluded whose projection is visualised.}
    \label{figure_se}   
\end{figure*}

Figures \ref{figure_isomap}-\ref{figure_se} shows visualization of DBM metric using 
Isometric Mapping (IsoMap) \cite{tenenbaum2000global},
t-Distributed Stochastic Neighbour Embedding (t-SNE) \cite{maaten2008visualizing}, 
Multi-dimensional Scaling (MDS) \cite{kruskal1964multidimensional}, and
Spectral Embedding (SE) \cite{belkin2003laplacian} 
respectively.
The characteristic of IsoMap is that it seeks a lower-dimensional embedding which maintains geodesic distances between all sample points that is it preserves the high-dimensional distance between the points.
% t-SNE converts affinities of data to probabilities. 
% Gaussian joint probabilities represent these affinities in the original space, and Student's t-distributions represent the affinities in the embedded space.
t-SNE tries to model similar data points in higher-dimensional space through small pairwise distances in lower-dimensional space.
In other words, it tries to minimise the Kullback–Leibler divergence between the two distributions of points in the map.
% t-SNE emphasises on; 
% (1) modelling dissimilar data points using large pairwise distances, and 
% (2) modelling similar data points using a small pairwise distance
% Our point of interest is that t-SNE tries to group the points in the similar cluster together, i.e. it tries to minimise the Kullback–Leibler divergence between the two distributions of points in the map.
% MDS seeks a lower-dimensional representation of the data in which the distances respect well the distances in the original high-dimensional space.
SE is a non-linear embedding, which finds a lower-dimensional representation of the sample points using a spectral decomposition of the graph Laplacian Eigenmaps. 
It is to be noted that 
Isomap (Figure \ref{figure_isomap}), 
t-SNE (Figure \ref{figure_tsne}), 
MDS (Figure \ref{figure_mds}), and 
SE (Figure \ref{figure_se}) 
are different visualisations for the same feature space $z$.
The idea for having these visualisations is to investigate whether the cluster for the unknown class can be segregated into one or more different classes.
In other words, we try to investigate whether there exists a single aggregate Amalgam Proportion for the unknown class or multiple 

The projections (Figures \ref{figure_tsne}-\ref{figure_se}) of CapsNet is uniform and dense while the other networks have more scattered non-uniform projections.
The non-uniform projection, which can be split into multiple clusters, of the other networks might suggest that the learned representation is not continuous/homogeneous enough. 
Interestingly, LeNet have more dense and uniform projections compared to other static neural networks, further suggesting the better representation of the LeNet.
These results are in accordance with the previous experiments (Section \ref{section_results}) on representation quality.

Another way to verify this interpretation is to look at the gaps in the projections, which is observing the behaviour of different data points. 
For CapsNet, even if we form clusters to have different classes, the gaps between the classes will be too small relative to other architectures. 
It also shows that in the high-dimensional space, all the soft-labels are moderately close to each other, also verified using Amalgam Metric (Table\ref{table_result} and Figure \ref{figure_histogram}).
While for the other architectures, there exist some points which can form their separate cluster and be termed as a different class.
Hence, for these architectures, it can have one or more different Amalgam proportion for the same unknown class which is contradicting to our hypothesis that there should exist only a single Amalgam proportion for a single unknown class. 
Note that, this dense projection does not necessarily mean that the unknown class has converged to a single known class.
It gives a visualisation that the Amalgam Proportion of the unknown class is similar.

\section{Visualisation Of Amalgam Metric (AM)}
\label{section_visualise_am}

\begin{figure*}[!htb]
    \centering
    \includegraphics[width=0.17\linewidth]{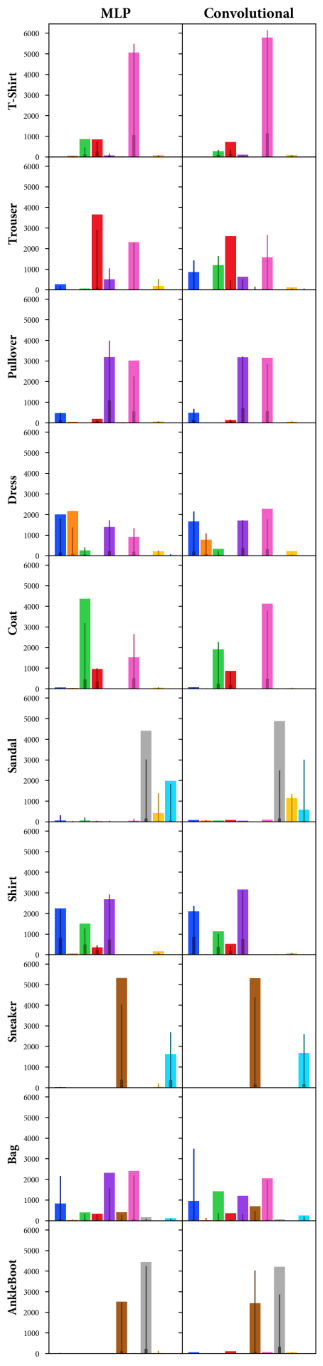} 
    \includegraphics[width=0.599\linewidth]{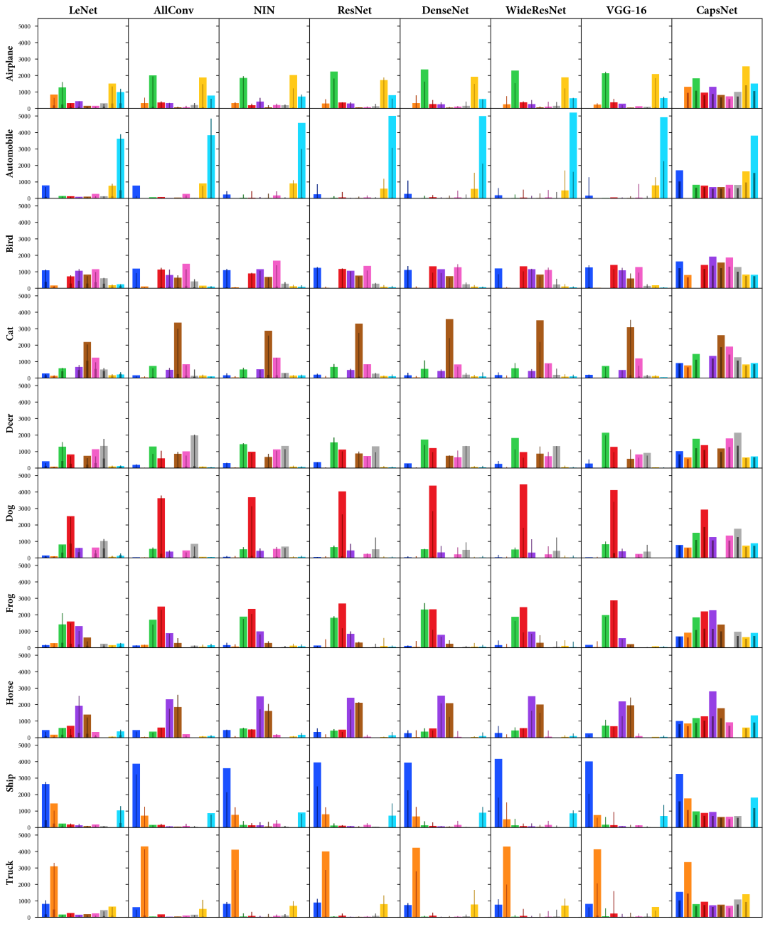}
    \includegraphics[width=0.17\linewidth]{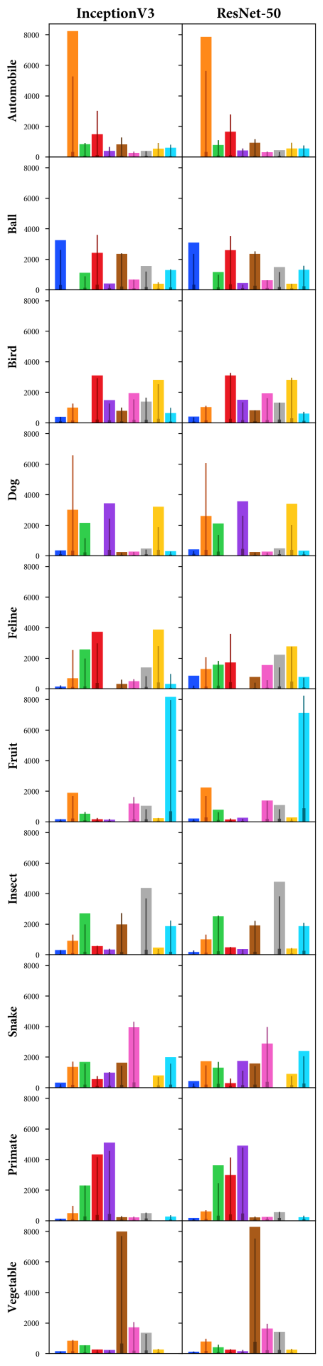} 
    \\
    Color Encoddings Of Classes For Fashion MNIST, CIFAR-10 and Sub Imagenet (In Order)\\
    \includegraphics[width=0.75\linewidth]{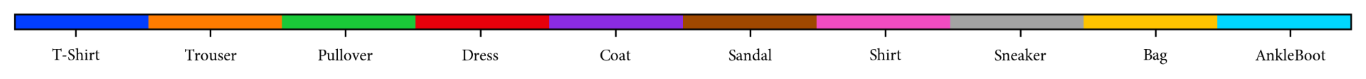} \\
    \includegraphics[width=0.75\linewidth]{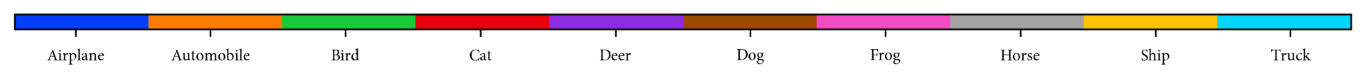}\\
    \includegraphics[width=0.75\linewidth]{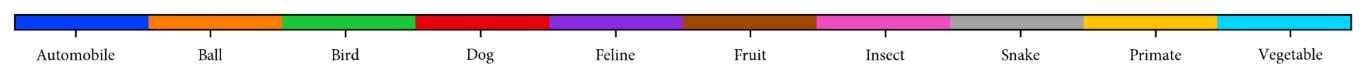} \\
    \caption{
    Histograms of soft-labels ($H'$ and $H$) from which the AM is calculated.
    Each row shows the histograms of one classifier with one class excluded.
    Dark-shaded thinner and light-shaded broader bins are respectively the soft-labels from the ground-truth $(H')$ from the classifier trained on all classes and the soft-labels of the classifier trained on $N-1$ classes $(H')$.
    }
    \label{figure_histogram}
\end{figure*}

To enable the visualisation of the Amalgam Metric, the computed histograms ($H$ and $H$) is plotted for every class and classifier (Figure \ref{figure_histogram}).
It is interesting to note that the histograms of CapsNet (Figure \ref{figure_histogram}) are different from the other ones, as architecture employed by CapsNet is entirely different.
This reveals that this metric can capture such representation differences.
It can be noted (Figure \ref{figure_histogram}) that for most classes of CapsNet, the variation is relatively low than the other architectures. 
This contributes to having a good representation of CapsNet. 

A further study can also be carried out to analyse the characteristics of representation of the neural network, which makes a class more robust than the other classes.
Further investigations can be also be carried out to analyse the effect of a class for an adversarial attack based on this. 
This can also provide insight into the classes which are robust to adversarial attacks. 
However, these analyses are out of scope for the current article, and hence, left for future work.  

\section{Visualisation Of Pearson Correlation}
\label{section_visualise_pearson}

\begin{figure*}[!htb]
    \centering
    \includegraphics[width=0.17\linewidth]{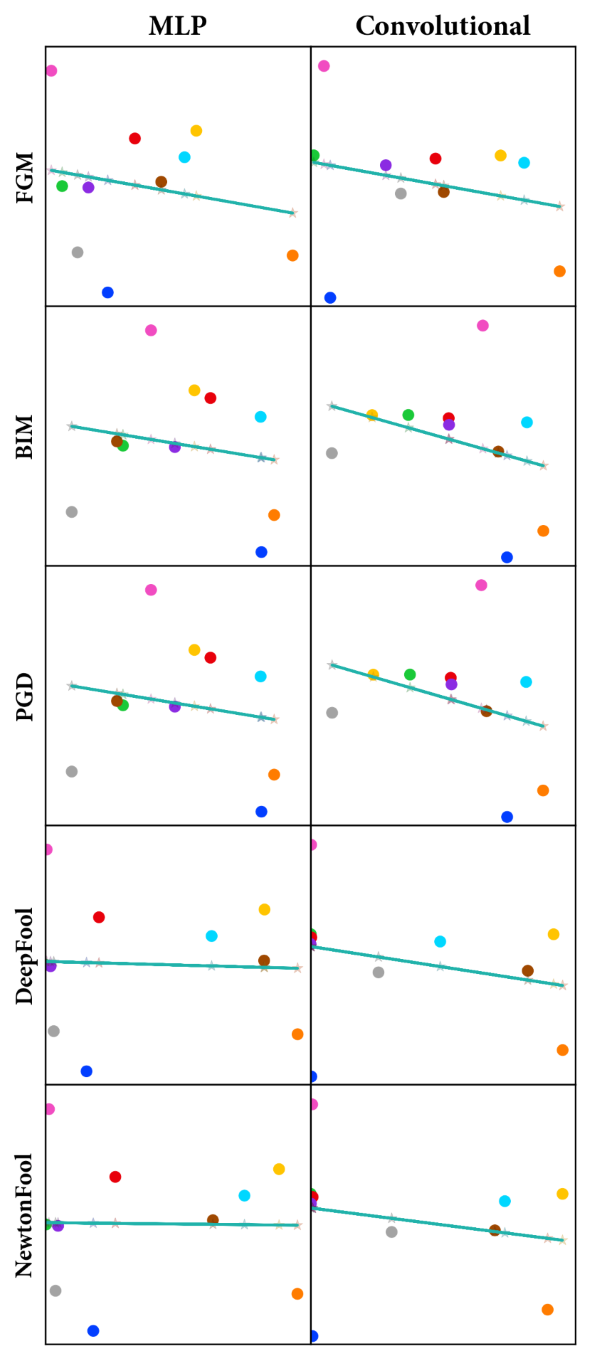} 
    \includegraphics[width=0.62\linewidth]{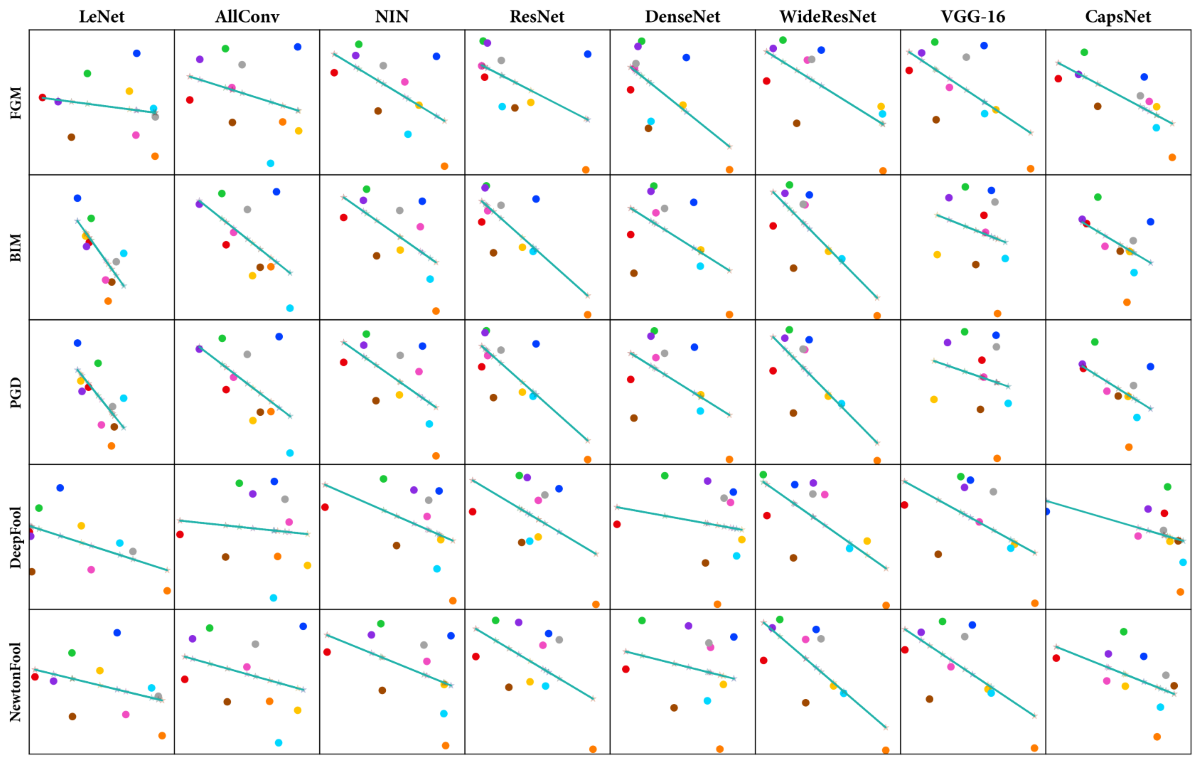}
    \includegraphics[width=0.17\linewidth]{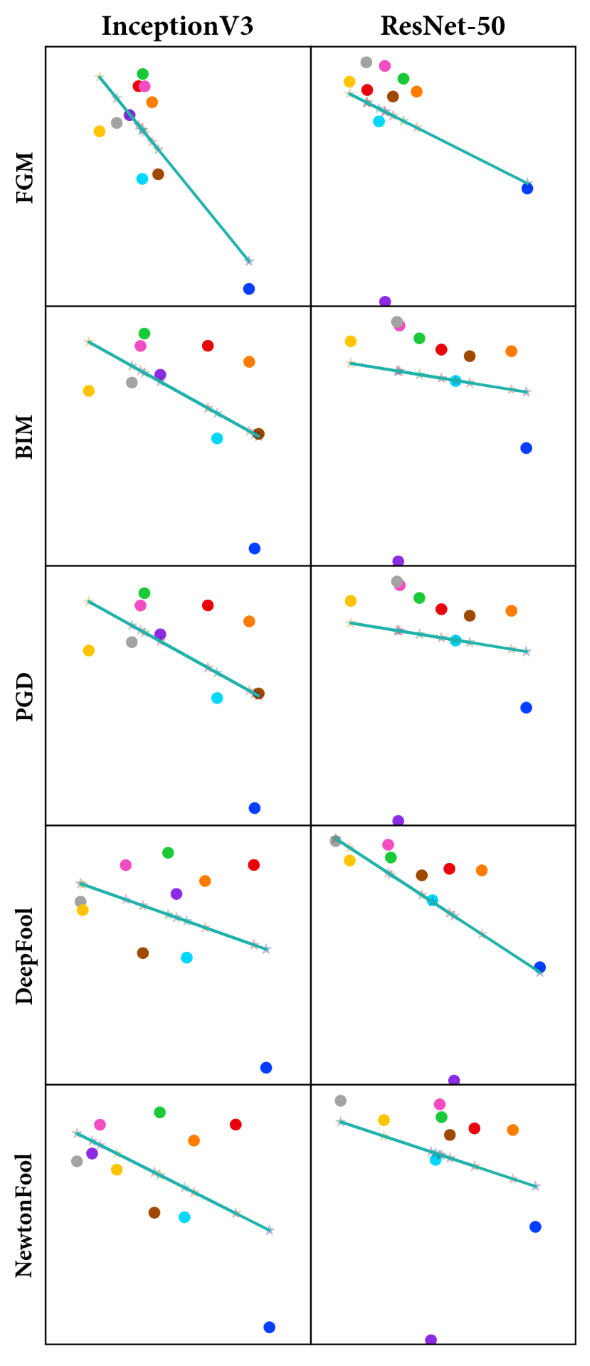} 
    \\
    Color Encoddings Of Classes For Fashion MNIST, CIFAR-10 and Sub Imagenet (In Order)\\
    \includegraphics[width=0.75\linewidth]{"images/FashionMnist/All/class_label".png} \\
    \includegraphics[width=0.75\linewidth]{"images/Cifar10/All/class_label".png}\\
    \includegraphics[width=0.75\linewidth]{"images/RestrictedImagenet/All/class_label".png} \\
    \caption{
        Visualisation of Pearson Correlation of Davies-Bouldin Metric (DBM) with Mean $L_2$ Score of adversarial attacks (Table \ref{table_pearson}). 
        Here, the x-axis represents the Mean $L_2$ Scores while the y-axis represents the DBM values and each point represent a DBM value and Mean $L_2$ Score for a labelled class.
    }
    \label{figure_pearson_dbm}
\end{figure*}

\begin{figure*}[!htb]
    \centering
    \includegraphics[width=0.17\linewidth]{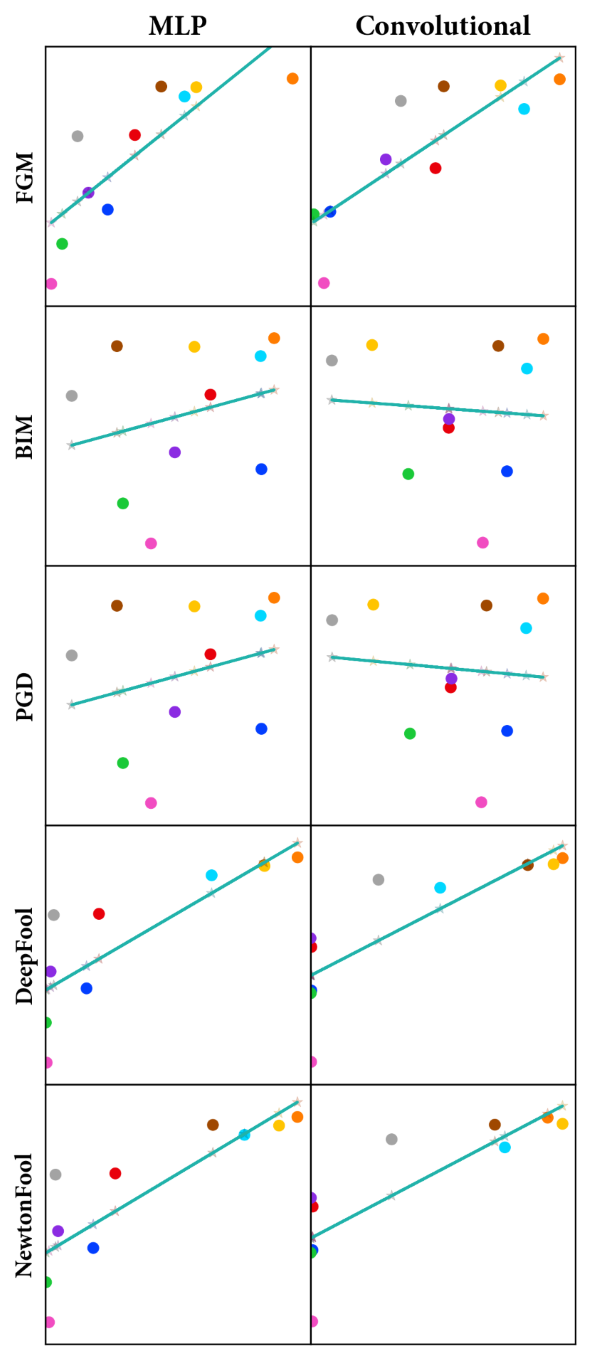} 
    \includegraphics[width=0.62\linewidth]{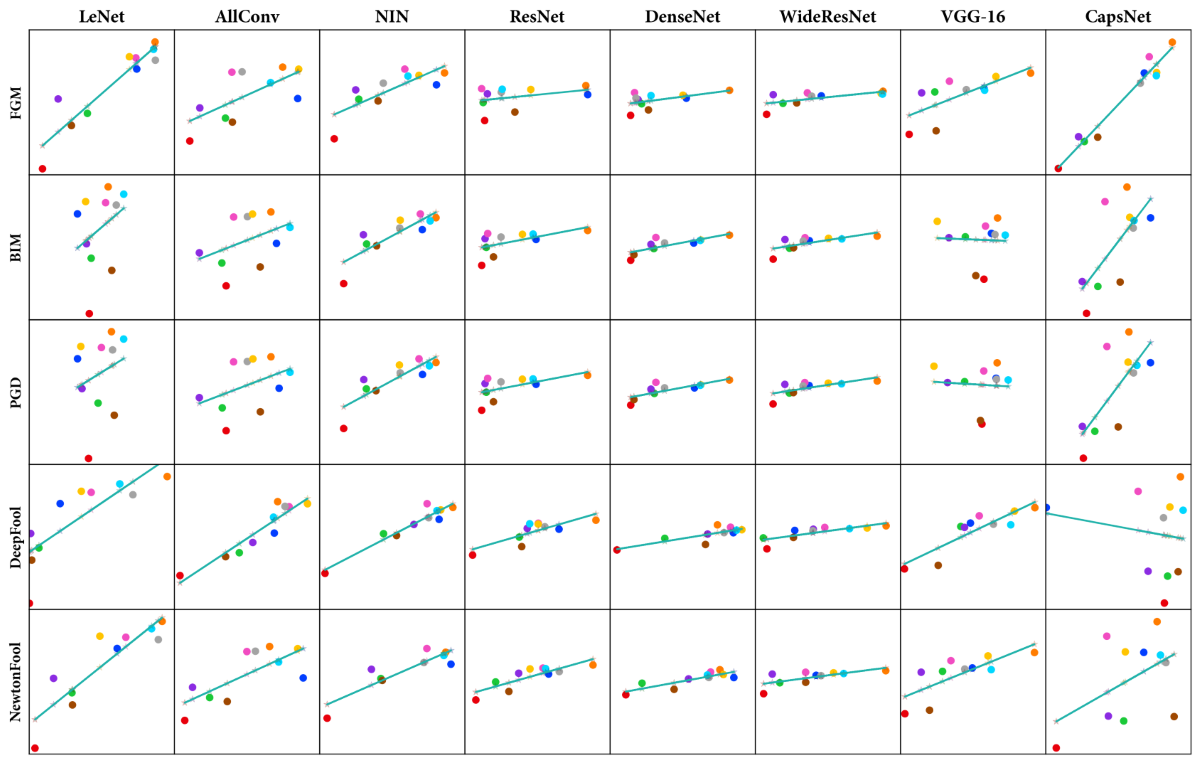}
    \includegraphics[width=0.17\linewidth]{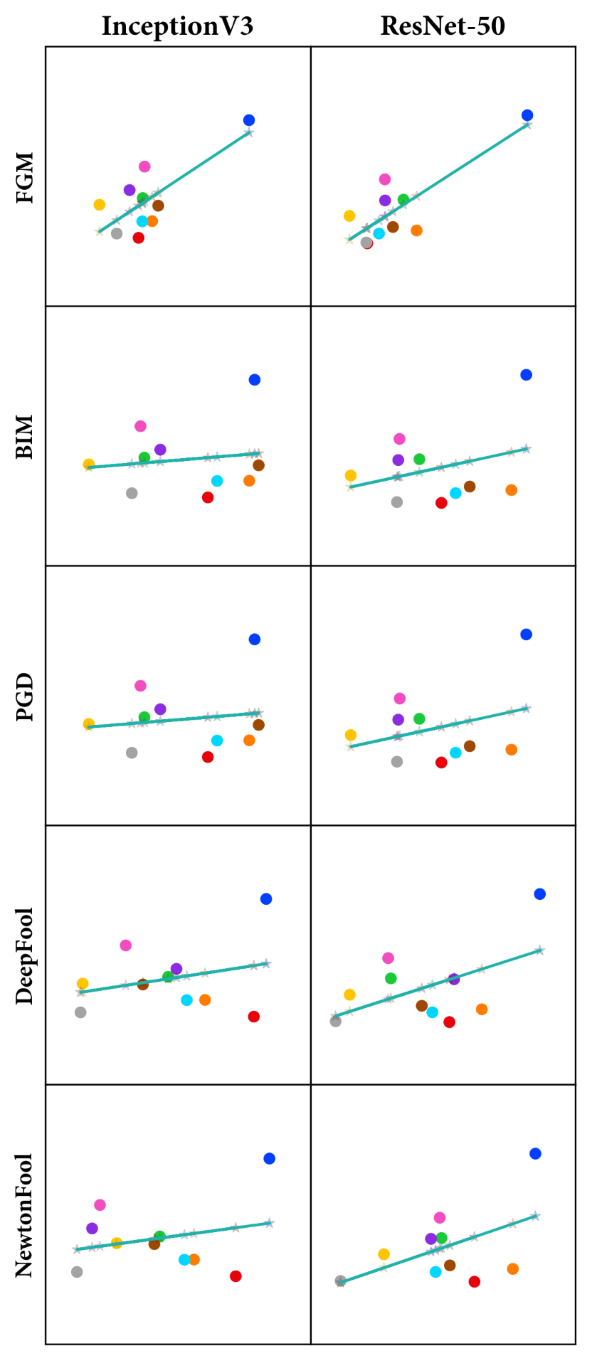} 
    \\
    Color Encoddings Of Classes For Fashion MNIST, CIFAR-10 and Sub Imagenet (In Order)\\
    \includegraphics[width=0.75\linewidth]{"images/FashionMnist/All/class_label".png} \\
    \includegraphics[width=0.75\linewidth]{"images/Cifar10/All/class_label".png}\\
    \includegraphics[width=0.75\linewidth]{"images/RestrictedImagenet/All/class_label".png} \\
    \caption{
        Visualisation of Pearson Correlation of Amalgam Metric (AM) with Mean $L_2$ Score of adversarial attacks (Table \ref{table_pearson}). 
        Here, the x-axis represents the Mean $L_2$ Scores while the y-axis represents the AM values and each point represent an AM value and Mean $L_2$ Score for a labelled class.
    }
    \label{figure_pearson_am}
\end{figure*}

Here, we visualise the Pearson correlation between the Raw Zero-Shot metrics (DBM and AM) with the adversarial metrics (Adversarial Accuracy and Mean $L_2$ Score) mentioned in Tables \ref{table_pearson}.
Figures  \ref{figure_pearson_dbm} and \ref{figure_pearson_am}, visualizes the relationship of Raw Zero-Shot metrics with adversarial metrics.
In the Section \ref{section_link_attack}, we observed some anomalies in the Pearson correlation values (Table \ref{table_pearson}).
Here we try to understand these anomalies with the help of our visualisation (Figure \ref{figure_pearson_am}).
On visualising the pearson correlation, we identify that DeepFool attacks the Airplane class of CapsNet with much less $L_2$ score compared to the other classes.
This abnormal behaviour of DeepFool for the Airplane class causes the anomaly for Pearson Correlation. 

\section{Another Outlook On Link Between Representation Quality And Adversarial Attacks}
\label{section_cdiff}

\begin{table*}[!htb]
    \centering
    \resizebox{0.95\linewidth}{!}{%
    \begin{tabular}{l|rrrrr|rrrrr}
        \toprule
        \multirow{2}{*}{\textbf{Classifier}} & \multicolumn{5}{c}{\textbf{Confidence Score}} & \multicolumn{5}{|c}{\textbf{Pearson Correlation of AM with Confidence Score}} \\
        & \multicolumn{1}{|c}{\textbf{FGM}} & \multicolumn{1}{c}{\textbf{BIM}} & \multicolumn{1}{c}{\textbf{PGD}} & \multicolumn{1}{c}{\textbf{DeepFool}} & \multicolumn{1}{c}{\textbf{NewtonFool}} 
        & \multicolumn{1}{|c}{\textbf{FGM}} & \multicolumn{1}{c}{\textbf{BIM}} & \multicolumn{1}{c}{\textbf{PGD}} & \multicolumn{1}{c}{\textbf{DeepFool}} & \multicolumn{1}{c}{\textbf{NewtonFool}} 
        \\
        \midrule
        \multicolumn{11}{c}{\textbf{Fashion MNIST}} \\
        \midrule 
        MLP     & 0.63 & 0.90 & 0.90 & 0.38     & 0.35       & 0.95 (0.00) & 0.99 (0.00) & 0.99 (0.00) & 0.86 (0.00) & 0.87 (0.00) \\
        ConvNet & 0.62 & 0.90 & 0.90 & 0.33     & 0.34       & 0.86 (0.00) & 0.99 (0.00) & 0.99 (0.00) & 0.85 (0.00) & 0.82 (0.00) \\
        \midrule
        \multicolumn{11}{c}{\textbf{CIFAR-10}} \\
        \midrule 
        LeNet      & 0.58 & 0.72 & 0.72 & 0.12     & 0.48    & 0.99 (0.00) & 1.00 (0.00) & 1.00 (0.00) & 0.79 (0.01)  & 1.00 (0.00)\\
        AllConv    & 0.82 & 0.91 & 0.91 & 0.71     & 0.69    & 0.92 (0.00) & 0.97 (0.00) & 0.97 (0.00) & 0.99 (0.00)  & 0.98 (0.00)\\
        NIN        & 0.87 & 0.93 & 0.93 & 0.78     & 0.75    & 0.94 (0.00) & 0.98 (0.00) & 0.98 (0.00) & 0.99 (0.00)  & 0.99 (0.00)\\
        ResNet     & 0.90 & 0.94 & 0.94 & 0.82     & 0.76    & 0.79 (0.01) & 0.91 (0.00) & 0.91 (0.00) & 0.96 (0.00)  & 0.94 (0.00)\\
        DenseNet   & 0.91 & 0.95 & 0.95 & 0.85     & 0.76    & 0.60 (0.07) & 0.95 (0.00) & 0.94 (0.00) & 0.91 (0.00)  & 0.94 (0.00)\\
        WideResNet & 0.87 & 0.97 & 0.97 & 0.84     & 0.77    & 0.31 (0.39) & 0.94 (0.00) & 0.94 (0.00) & 0.96 (0.00)  & 0.65 (0.04)\\
        VGG-16     & 0.86 & 0.95 & 0.95 & 0.82     & 0.75    & 0.89 (0.00) & 0.98 (0.00) & 0.98 (0.00) & 0.93 (0.00)  & 0.95 (0.00)\\
        CapsNet    & 0.17 & 0.46 & 0.48 & -0.10    & 0.15    & 0.89 (0.00) & 0.93 (0.00) & 0.93 (0.00) & -0.52 (0.13) & 0.24 (0.50)\\
        \midrule 
        \multicolumn{11}{c}{\textbf{Sub-Imagenet}} \\
        \midrule
        InceptionV3 & 0.92 & 0.95 & 0.95 & 0.86     & 0.82       & 0.10 (0.78) & 0.54 (0.11) & 0.54 (0.11) & 0.33 (0.35) & 0.14 (0.70) \\
        ResNet-50   & 0.90 & 0.94 & 0.94 & 0.85     & 0.81       & 0.44 (0.20) & 0.75 (0.01) & 0.75 (0.01) & 0.67 (0.03) & 0.53 (0.12) \\
        \bottomrule
    \end{tabular}
    }
    \caption{Confidence Difference Score and it's Pearson Correation value (and p-value) for each classifier and adversarial attack pair.}
    \label{table_cdiff}
\end{table*}

\begin{figure*}[!htb]
    \centering
    \includegraphics[width=0.17\linewidth]{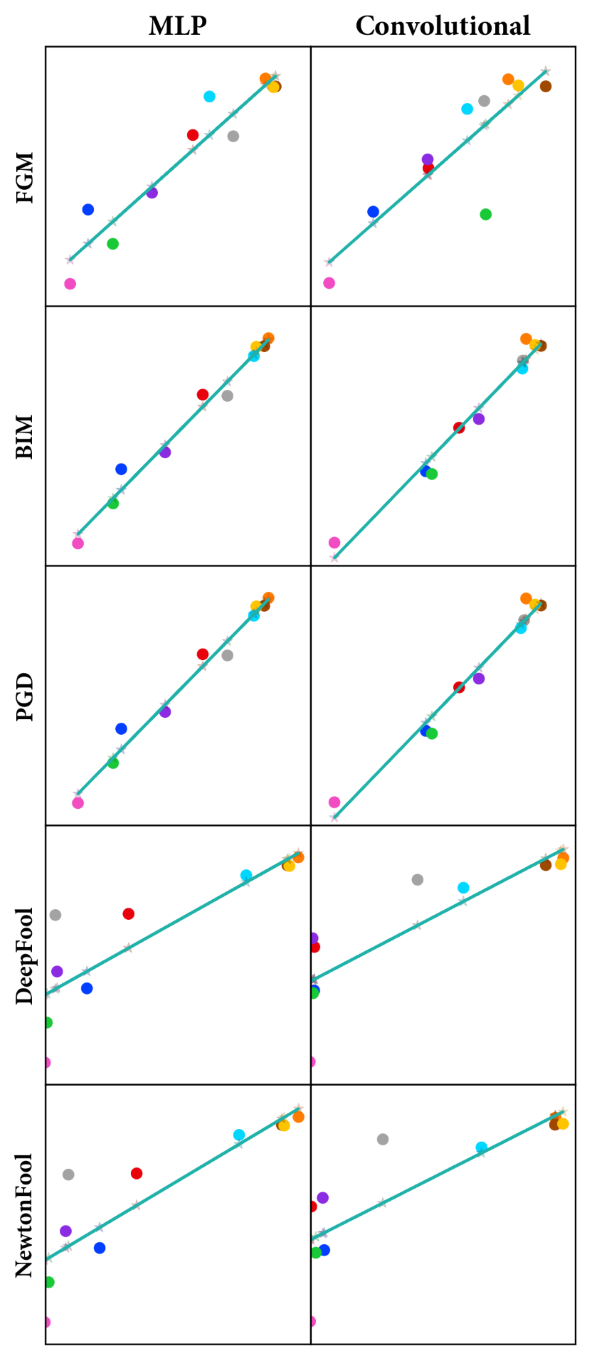} 
    \includegraphics[width=0.62\linewidth]{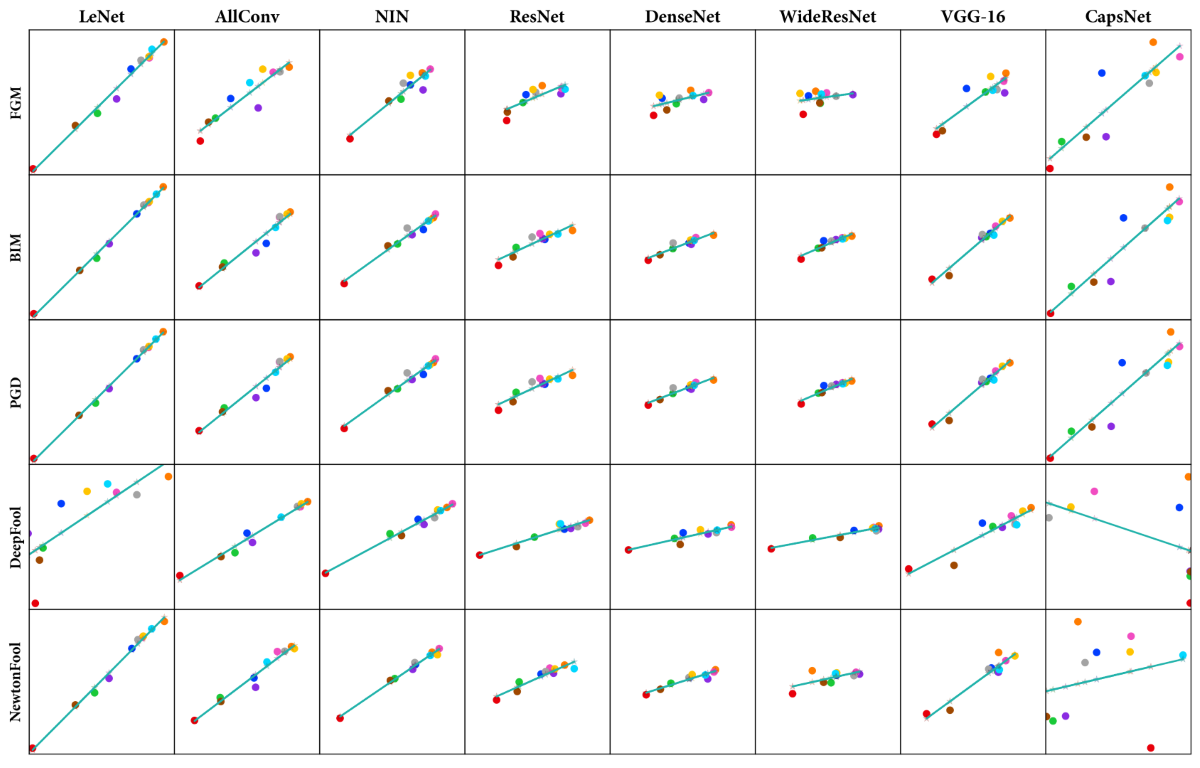}
    \includegraphics[width=0.17\linewidth]{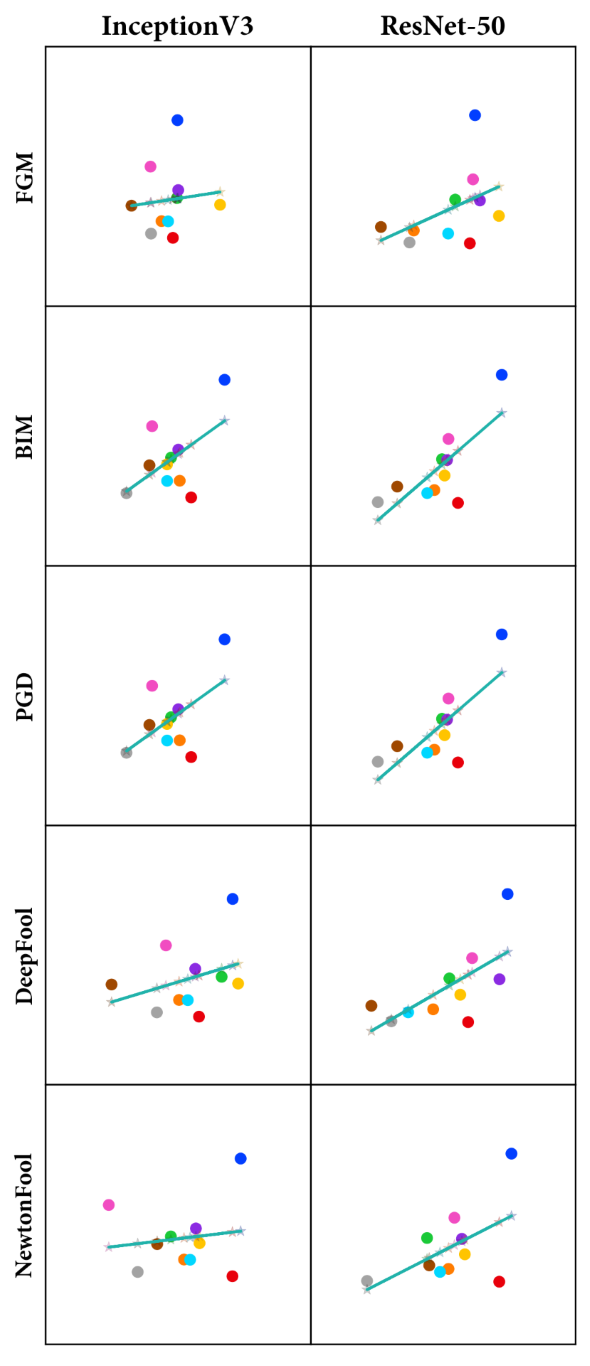} 
    \\
    Color Encoddings Of Classes For Fashion MNIST, CIFAR-10 and Sub Imagenet (In Order)\\
    \includegraphics[width=0.75\linewidth]{"images/FashionMnist/All/class_label".png} \\
    \includegraphics[width=0.75\linewidth]{"images/Cifar10/All/class_label".png}\\
    \includegraphics[width=0.75\linewidth]{"images/RestrictedImagenet/All/class_label".png} \\
    \caption{
        Visualisation of Pearson Correlation of Amalgam Metric (AM) with Confidence Score of adversarial attacks (Table \ref{table_cdiff}). 
        Here, the x-axis represents the Confidence Difference Scores. In contrast, the y-axis represents the AM values, and each point represents an AM value and Confidence Difference Scores for a labelled class.
    }
    \label{figure_pearson_cdiff}
\end{figure*}

In this section, we analyse the representation quality from the perspective of Confidence Score, which is defined as the change in the confidence of the true label by an adversarial sample. 
To further deeply analyse the statistical relevance of this link between representation quality and adversarial attacks, we here conduct a Pearson Correlation test of Amalgam Metric of the vanilla classifiers with Confidence Score of the adversarial attacks. 
The Pearson correlation value of the Amalgam Metric with Confidence Score is shown in Table \ref{table_cdiff} for every architecture and attacks. 
Table \ref{table_cdiff} also mentions the Confidence Score of every classifier-attack pair. 
Moreover, similar to our previous analysis, these Pearson relationships between the Amalgam Metric and Confidence Score can also be visualised (Figure \ref{figure_pearson_cdiff}). 

The purpose of evaluating the Confidence Score as an adversarial metric is because the score effectively assesses the impact of the adversarial attacks on true class soft-label.
This perspective gives us the effectiveness of an attack on the soft-label of the representation, we evaluate. 
% As some of the previous researches have claimed, that many of the adversarial attacks force the misclassification by 
Therefore, here Confidence Score not only determines the alteration in the representation space, but it also analyses the effectiveness of an attack across different classes.

The correlational analysis of our Amalgam Metric suggests a strong relationship between our Amalgam Metric and the adversarial attacks in general.
We do observe some anomalies in this Pearson correlation also with AM of DeepFool for CapsNet.
However, we believe this anomaly is due to the adversarial attack itself. 
Note that in Table \ref{table_cdiff} the Confidence Score of the DeepFool attack for CapsNet is negative, which suggests that DeepFool instead of decreasing the soft-label of the true-class, increases the soft-label of the misclassified class.
We do note that, more investigations are required to better understand the behaviour of Capsule Networks, in general.

\end{document}